\documentclass{article}
\usepackage[accepted]{icml2020}
\usepackage{url,amsthm,amsmath,amssymb,amsfonts}
\usepackage{xcolor}
\usepackage{algpseudocode}
\usepackage{tikz}
\usepackage{bbm}
\usepackage{rbase} 
\usepackage{paralist}
\usepackage{enumitem}
\definecolor{hcitecolor}{RGB}{40,40,160}
\usepackage[colorlinks=true,citecolor=hcitecolor, linkcolor=black]{hyperref}
\usepackage{graphicx}
\usepackage{pgfplots}
\usepackage{xypic}
\usepackage[all,2cell,ps,cmtip]{xy}
\usepackage{tikz}
\usepackage{caption}
\usepackage{microtype}
\usepackage{mdframed}
\usepackage{young}
\usepackage[enableskew]{youngtab}
\Yboxdim{8pt}
\usepackage{ifthen}
\newcommand{\ifdraft}[1]{}
\newcommand{\ificml}[1]{#1}

\algdef{SE}[DOWHILE]{Do}{doWhile}{\algorithmicdo}[1]{\algorithmicwhile\ #1}%
\algnewcommand{\varalg}{\textit}

\usepackage{tikz}
\usetikzlibrary{shadows}
\usetikzlibrary{matrix, calc, positioning, arrows,shapes,backgrounds, arrows.meta, quotes}
\usepackage{gensymb,stmaryrd,mathtools,textcomp,xspace,grffile,rbase,soul}

\newtheorem{theorem}{Theorem}

\newtheorem{prop}[theorem]{Proposition}

\newtheorem{corollary}[theorem]{Corollary}

\newtheorem{definition}{Definition}

\newenvironment{pf}{\textbf{Proof.}}{\mbox{}\hfill\m{\blacksquare}\\} % \mbox{}\noindent}
\newenvironment{pfof}[1]{\textbf{Proof of #1.}}{\mbox{}\hfill\m{\blacksquare}\\ \mbox{}\noindent}

\ignore{
\makeatletter
\def\thm@space@setup{%
  \thm@preskip=6pt plus 0pt minus 0pt%5cm plus 1cm minus 2cm
  \thm@postskip=0pt plus 0pt minus 0pt %\thm@preskip % or whatever, if you don't want them to be equal
}
\makeatother
}

\newcommand{\includepicc}[2]{\centerline{\includegraphics[width=#1\textwidth]{#2}}}

\newcommand{\neuron}{\mathfrak{n}}
\newcommand{\actn}[1]{f^{#1}}
\newcommand{\actf}[1]{\mathfrak{f}_{#1}}
\newcommand{\fout}{f^{out}}
\newcommand{\ffin}{\mathfrak{f}^{\textrm{in}}}
\newcommand{\ffout}{\mathfrak{f}^{\textrm{out}}}
\newcommand{\fin}{f^{\textrm{in}}}
\newcommand{\finb}{\bar{f}}
\newcommand{\fpre}{\tilde{f}}
\newcommand{\inpx}{x}
\newcommand{\outy}{y}

\newcommand{\Snm}{\Sbb_{n-1}}
\newcommand{\Snmm}{\Sbb_{n-2}}
\newcommand{\domr}{\mathop{\mathrm{dom}}}

\begin{document}

\title{The general theory of permutation equivarant neural networks and higher order 
graph variational encoders}
\twocolumn[
\icmltitle{The general theory of permutation equivarant neural networks and higher order
graph variational encoders}
\begin{icmlauthorlist}
\icmlauthor{Erik Henning Thiede}{UofC,flatiron}
\icmlauthor{Truong Son Hy}{UofC}
\icmlauthor{Risi Kondor}{flatiron,UofC}
\end{icmlauthorlist}
\vskip 10pt 

\icmlaffiliation{UofC}{Department of Computer Science, University of Chicago, Chicago, Illinois, USA}
\icmlaffiliation{flatiron}{Flatiron Institute, New York City, New York, USA}

\icmlcorrespondingauthor{Erik Henning Thiede}{thiede@uchicago.edu}
\icmlcorrespondingauthor{Risi Kondor}{rkondor@flatironinstitute.org}

% \centerline{\textsc{Paper under review. Please do not distribute.}}
\vskip 0.3in
]

\printAffiliationsNoNotice{}
% \printAffiliationsAndNotice{}

\begin{abstract}
Previous work on symmetric group equivariant neural networks generally only considered the 
case where the group acts by permuting the elements of a single vector. 
In this paper we derive formulae for general permutation equivariant layers, including 
the case where the layer acts on matrices by permuting their rows and columns simultaneously. 
This case arises naturally in graph learning and relation learning applications. 
As a specific case of higher order permutation equivariant networks, we present a second order 
graph variational encoder, and show that the latent distribution of equivariant generative models must be 
exchangeable.  
We demonstrate the efficacy of this architecture on the tasks of link prediction in citation graphs and molecular graph generation.

\end{abstract}

\section{Introduction}

%Behavior under symmetries is a crucial feature of many learning algorithms. 
%In the context of multilayer neural networks this behavior is 

Generalizing from the success of convolutional neural networks in computer vision, 
equivariance has emerged as a core organzing principle of deep neural network architectures. 
%Equivariance to a group of transformations \m{G} means 
%that if the inputs of the network are transformed by some group element \m{g\tin G}, 
%then the output of each subsequent layer \m{\ell}, regarded as a single vector \sm{\actn{\ell}}, must  
%transform to \sm{T^\ell_g(\actn{\ell})} for some fixed set of linear transformations 
%\sm{\setofN{\ts T^\ell_g\ts }{g\tin G\ts}}. %It is easy to see that in any 
%The classical example of equivariant networks are the convolutional neural nets (CNNs) 
%used in image recognition, which are 
Classical CNNs are equivariant to translations  
%in which case \m{G} is the group of translations of the input images, and for each translation \m{g}, 
%\sm{T^\ell_g} is the corresponding translation of the activations in layer \m{\ell} 
\cite{LeCun1989}. 
In recent years, starting with \cite{Cohen2016}, 
researchers have also constructed networks that are equivariant to 
the three dimensional rotation group 
\cite{CohenSpherical2018} 
\cite{CGnetsNIPS2018}, 
the Euclidean group of translations and rotations \cite{Cohen2017}\cite{Weiler2018arxiv}, 
and other symmetry groups \cite{Poczos2017}. 
Closely related are generalizations of convolution to manifolds 
\cite{Marcos2017}\cite{Worrall2017}. 
Gauge equivariant CNNs form an overarching framework that connects the two domains \cite{CohenGaugeICML2019}.

The set of all permutations of \m{n} objects also forms a group, called the \emph{symmetric group} 
of degree \m{n}, commonly denoted \m{\Sn}. 
The concept of permutation equivariant neural networks was proposed in \cite{Guttenberg16},  
and discussed in depth in ``Deep sets'' by \cite{DeepSets}. 
Since then, permutation equivariant models have found applications in a number of domains, including 
understanding the compositional structure of language \cite{gordon2020iclr}, 
and such models were analyzed from theoretical point of view in \cite{KerivenPeyre2019}
\cite{Sannai2019universal}. 
%equivariance was discussed in XXX, XXX and XXX. 
The common feature of all of these approaches however is that they only consider one specific way that 
permutations can act on vectors, namely 
\sm{(f_1,\ldots,f_n)\mapsto(f_{\sigma^{-1}(1)},\ldots,f_{\sigma^{-1}(n)})}. 
This is not sufficient to describe certain naturally occurring situations, for example, when \m{\Sn} 
permutes the rows and columns of an adjacency matrix. 

In the present paper we generalize the notion of permutation equivariant neural networks to other 
actions of the symmetric group, and derive the explicit form of the corresponding equivariant layers, 
including how many learnable parameters they can have. 
In this sense our paper is similar to recent works such as \cite{CohenCNNhomo,KondorTrivedi2018,Yarotsky2018universal} 
which examined the algebric aspects of equivariant nets, but with a specific focus on the symmetric group.   
%In particular, we discuss the second order permutation action that 
%describes how permutations act on the adjacency matrix of a graph, or more generally, how 
%permutations transform a matrix describing any binary relation. 

On the practical side, higher order permutation equivariant neural networks appear naturally 
in graph learning and graph generation \cite{maron2018invariant,CCN-JCP}. More generally, 
we argue that this symmetry is critical for encoding relations between pairs, triples, quadruples etc.~of entities rather 
than just whether a given object is a member of a set or not. 
As a specific example of our framework we present a second order equivariant graph variational encoder and 
demonstrate its use on link preduction and graph generation tasks. 
The distribution on the latent layer of such a model must be exchangeable (but not necessarily IID), forming 
and interesting connection to Bayesian nonparametric models \cite{BloemReddy2019}. 

\section{Equivariance to permutations}

A permutation (of order \m{n}) is a bijective map \m{\sigma\colon \cbrN{\oneton{n}}\to\cbrN{\oneton{n}}}. 
The product of one permutation \m{\sigma_1} with another permutation 
\m{\sigma_2} is the permutation that we get by first performing \m{\sigma_1}, then \m{\sigma_2}, i.e., 
%defined as the permutation corresponding to composing the two maps, i.e., 
\m{(\sigma_2\sigma_1)(i):=\sigma_2(\sigma_1(i))}. %, then with respect to this operation 
It is easy to see that with respect to this notion of product, 
the set of all \m{n!} permutations of order \m{n} form a \emph{group}. %As mentioned in the Introduction, 
This group is called the \emph{symmetric group of degree} \m{n}, and denoted \m{\Sn}. 

Now consider a feed-forward neural network consisting of \m{s} neurons, \m{\seq{\neuron}{s}}. 
We will denote the activation of the \m{i}'th neuron \m{\actn{i}}. 
Each activation may be a scalar, a vector, a matrix or a tensor. 
% (\sm{\actn{i}\tin\RR}) or a vector (\sm{\actn{i}\tin \RR^{D_i}}). 
%In this paper we are interested in the case when the inputs to our network are 
As usual, we assume that the input to our network is a fixed size 
vector/matrix/tensor \m{\inpx}, and the ouput is a fixed sized vector/matrix/tensor \m{\outy}.   
%vectors \sm{x\tin \RR^{D_{\textrm{in}}}}, and the output is a fixed-length vector 
%\sm{y\tin \RR^{D_{\textrm{out}}}}.

The focus of the present paper is to study the behavior of neural networks under the 
action of \m{\Sn} on the input \m{\inpx}. This encompasses a range of special cases, relevant 
to different applications:	
%\begin{compactenum}[~~~1.]
%\setlength{\partopsep}{0pt}
%\setlength{\pltopsep}{0pt}
%\setlength{\itemsep}{0pt}
\begin{enumerate}[itemsep=0pt, topsep=0pt, leftmargin=12pt, labelsep=4pt, label=\arabic*.]
\item \textbf{Trivial action.} The simplest case is when \m{\Sn} acts on \m{\inpx} trivially, i.e., 
\m{\sigma(\inpx)\<=\inpx}, so permutations don't change \m{\inpx} at all. 
This case is not very interesting for our purposes. 
\item \textbf{First order permutation action.} The simplest non-trivial \m{\Sn}--action is when 
\m{\inpx} is an \m{n} dimensional vector and \m{\Sn} permutes its elements: 
\begin{equation}\label{eq: first order action}
[\sigma(\inpx)]_i=\inpx_{\sigma^{-1}(i)}.
\end{equation}
This is the case that was investigated in \citep{DeepSets}  
because it arises naturally when learning from \emph{sets}, 
in particular when \m{\inpx_i} relates to the \m{i}'th element of a set \m{S} of \m{n} objects \cbrN{\sseq{o}{n}}. 
Permuting the numbering of the objects does not change \m{S} as a set, but it does change 
the ordering of the elements of \m{\inpx} exactly as in \rf{eq: first order action}. 
When learning from sets the goal is to 
construct a network which, as a whole, is \emph{invariant} to the permutation action. 
A natural extension allowing us to describe each object with more than just a 
single number is when \m{\inpx} is an \m{n\times d} dimensional \emph{matrix} on which \m{\Sn} acts  
by permuting its rows, \m{[\sigma(\inpx)]_{i,j}=\inpx_{\sigma^{-1}(i),\,j}}. 
\item \textbf{Second order permutation action.} 
The second level in the hierarchy of permutation actions 
is the case when \m{\inpx} is an \m{n\<\times n} matrix 
on which the symmetric group acts by permuting both its rows and its columns:
\[[\sigma(\inpx)]_{i,j}=\inpx_{\sigma^{-1}(i),\,\sigma^{-1}(j)}.\] 
While this might look exotic at first sight, it is exactly the case faced by graph neural networks, 
where \m{\inpx} is the adjacency matrix of a graph. 
More generally, this case encompasses any situation involving learning from binary 
relations on a set. 
%In graph neural networks the objective is to be invariant to this action. 
%In other applications, however, we will want the network to be equivariant rather than 
%invariant (see below). 
\item \textbf{Higher order cases.} 
%The natural generalization of the above is when \m{x} is a tensor of 
Extending the above, if \m{x} is a tensor of order \m{k}, the \m{k}'th order permutation action of 
\m{\Sn} on \m{x} transforms it as 
\begin{equation}\label{eq: kth order action}
[\sigma(\inpx)]_{\sseq{i}{k}}=\inpx_{\sigma^{-1}(i_1),\ldots,\sigma^{-1}(i_k)}.
\end{equation}
This case arises, for example, in problems involving rankings, and was also investigated in \cite{maron2018invariant}.
\item \textbf{Other actions.} 
Not all actions of \m{\Sn} can be reduced to actually permuting the elements of a tensor. 
We will discuss more general cases in Section \ref{sec: fourier}. 
%Higher order cases and other actions of \m{\Sn} 
%arise for example when using neural networks to learn from permuations. 
%We defer discussing this to Section X.   
\end{enumerate}
\begin{figure}
\includepicc{.35}{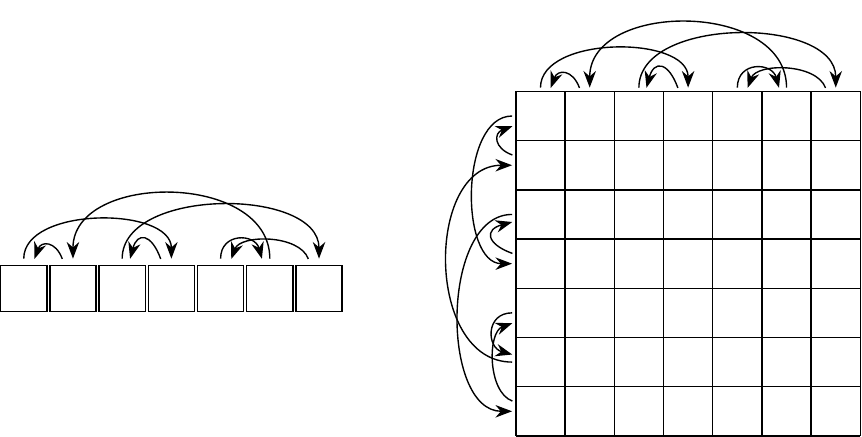}
%\begin{center}
%\begin{minipage}{.5\textwidth}
\caption{The symmetric group acts on vectors by permuting their elements (left). 
However, in this paper we also consider equivariance to other types of \m{\Sn}--actions, such 
as the way that a single permutation \m{\sigma\tin\Sn} 
permutes the rows and columns of an adjacency matrix simultaneously (right). 
The most general types of \m{\Sn}--actions are best expressed in Fourier space. 
\label{fig: actions}}
%\end{minipage}
\vspace{-0.2in}
%\end{center}
\end{figure}

\subsection{Invariance vs.~equivariance}

Neural networks learning from sets or graphs must be invariant 
to the action of the symmetric group on their inputs. 
However, even when a network is overall invariant, its internal activations 
%The internal activations of the network, however 
are often expected to be equivariant (covariant) rather than invariant. 
In graph neural networks, for example, the output of the \m{\ell}'th layer   
is often a matrix \sm{\actn{\ell}} whose rows are indexed by the vertices \citep{BrunaZaremba2014}. 
If we permute the vertices, %the rows of \sm{f_\ell} correspondingly permute, 
\sm{\actn{\ell}} will to change to 
$\actn{\ell}\prime$, 
where \sm{[\actn{\ell}\prime]_{i,j}=[\actn{\ell}]_{\sigma^{-1}(i),\,j}}. 
It is only at the top of the network that invariance is enforced, typically by summing over 
the \m{i} index of the penultimate layer.

Other applications demand that the output of the network be \emph{covariant} with the permutation 
action. Consider, for example the case of learning to fill in the missing edges of a graph of 
\m{n} vertices. In this case both the inputs and the outputs of the network are \m{n\<\times n} 
adjacency matrices, so the output must transform according to the same action as the input. 
%\[\outy\mapsto \outy'\qqquad\qqquad \outy'_{i,j}=\outy_{\sigma^{-1}(i),\,\sigma^{-1}(j)}.\] 
Naturally, the internal nodes of such a network must also co-vary with permutations. 
action on the inputs and cannot just be invariant.

%For generality, 
In this paper we assume that every activation of the network  
covaries with permutations. However, 
%the way that each activation changes under permutations may be different, i.e., 
each \m{\actn{i}} may transform according to a different action of the symmetric group. 
The general term for how these activations transform in a coordinated way is \emph{equivariance}, which 
we define formally below. 
Note that invariance is a special case of equivariance corresponding to the trivial \m{\Sn}--action 
\sm{\sigma(\actn{i})\<=\actn{i}}. 
Finally, we note that in this paper we use the terms \emph{covariant} and \emph{equivariant} essentially 
interchangeably: in general, the former is more commonly used in the context of compositional architectures, 
whereas the latter is the generic term used for convolutional nets. 

\subsection{General definition of permutation equivariance}

%All these different cases demand that we provide a rather general definition of equivariance to 
%permutati
%To unify all the above cases and facilitate our further discussion 
%we need a rather general definition of equivariance to permutations.
%We start with a formal definition of symmetric group actions. 

%Before defining \m{\Sn}--equivariance, 
To keep our discussion as general as possible, we start with a general definition of symmetric group actions. 
\ifdraft{\smallskip} 

\begin{definition}
Let \m{f} be a vector, matrix or tensor representing the input to a neural network or 
the activation of one of its neurons. We say that the symmetric group \m{\Sn} \df{acts linearly}  
on \m{f} if under a permutation \m{\sigma\tin\Sn}, \m{f} changes to \m{T_\sigma(f)} 
for some fixed collection of linear maps \sm{\setofN{T_\sigma}{\sigma\tin\Sn}}. 
\end{definition}

This definition is more general than the first, second and \m{k}'th order 
permutation actions described above, because it does 
not constrain \m{T_\sigma} to just permute the entries of \m{f}. Rather, \m{T_\sigma} can 
be any linear map. % that behaves \emph{like} a permutation. 
Our general notion of permutation equivariant networks is then the following. 
\ifdraft{\smallskip}

\begin{definition}\label{eq: equivariance}
Let \m{\Ncal} be a neural network whose input is \m{x} and whose activations are \m{\actn{1},\actn{2},\ldots \actn{s}}. 
Assume that the symmetric group \m{\Sn} acts on \m{x} linearly by the maps 
\sm{\setofN{T^{\textrm{in}}_\sigma}{\sigma\tin\Sn}}. 
%under a permutation \m{\sigma\tin\Sn} the inputs of \m{\Ncal} change as 
%\sm{x\mapsto T^{\textrm{in}}_\sigma(x)} 
%for some fixed set of linear maps \sm{\setofN{T^{\textrm{in}}_\sigma}{\sigma\tin\Sn}}. 
We say that \m{\Ncal} is \df{equivariant to permutations} if each of its neurons has a 
corresponding \m{\Sn}--action \sm{\setofN{T^{i}_\sigma}{\sigma\tin\Sn}} 
such that when \m{x\mapsto T^{\textrm{in}}_\sigma(x)},~\mbox{} \m{\actn{i}} will correspondingly transform as  
\sm{\actn{i}\mapsto T^{i}_\sigma(\actn{i})}. 
\end{definition} 

Note that \citep{DeepSets} only considered the case of first order permutation actions  
%defined permutation equivariance in terms of only the first order permutation action 
\begin{equation}\label{eq: first order}
\ifdraft{
\brN{[\actn{i}]_1,\ldots,[\actn{i}]_n}\overset{\sigma}{\longmapsto}  
\brN{[\actn{i}]_{\sigma^{-1}(1)},\ldots,[\actn{i}]_{\sigma^{-1}(n)}}.
}
\ificml{
\brN{\actn{i}_1,\ldots,\actn{i}_n}\overset{\sigma}{\longmapsto}  
\brN{\actn{i}_{\sigma^{-1}(1)},\ldots,\actn{i}_{\sigma^{-1}(n)}}, 
}
\end{equation}
whereas our definition % is considerably more general and allows us to discuss 
also covers richer forms of equivariance, including the second order case 
\ifdraft{  
\begin{equation}\label{eq: second order}
\brN{[\actn{i}]_{1,1},[\actn{i}]_{1,2},\ldots,[\actn{i}]_{n,n}}\overset{\sigma}{\longmapsto}  
\brN{[\actn{i}]_{\sigma^{-1}(1),\sigma^{-1}(1)},[\actn{i}]_{\sigma^{-1}(1),\sigma^{-1}(2)},\ldots,
[\actn{i}]_{\sigma^{-1}(n),\sigma^{-1}(n)}}
\end{equation}}
\ificml{\vspace{-6pt}  
\begin{multline}\label{eq: second order}
\brN{\actn{i}_{1,1},\actn{i}_{1,2},\ldots,\actn{i}_{n,n}}\overset{\sigma}{\longmapsto}\\  
\brN{\actn{i}_{\sigma^{-1}(1),\sigma^{-1}(1)},\actn{i}_{\sigma^{-1}(1),\sigma^{-1}(2)},\ldots,
\actn{i}_{\sigma^{-1}(n),\sigma^{-1}(n)}}
\end{multline}
}
that is relevant to graph and learning relation learning. % applications.  
 %Let \m{\Ncal} be a feed-forward neural network in which the neurons are arranged in \m{L\<+1} 
%layers labeled \m{\ell=0,1,\ldots L}. Collectively, the activations of all neurons in 

%Consider a feed-forward neural network consisting of \m{s} neurons \m{\seq{\neuron}{s}}. 
%Assume that the activation of each neuron \m{\neuron_i} is a scalar \sm{f_{\neuron_i}\tin\RR} or 
%a vector \sm{f_{\neuron_i}\tin \RR^{D_i}}. 
%In this paper we are interested in the case when the inputs to our network are 
%As usual, we also assume that the inputs to our network are fixed length 
%vectors \sm{x\tin \RR^{D_{\textrm{in}}}}, and the output is a fixed-length vector 
%\sm{y\tin \RR^{D_{\textrm{out}}}}.

\section{Convolutional architectures}

Classical convolutional networks and their generalizations to groups 
are characterized by the following three features:
%\begin{compactenum}[~~~1.]
\begin{enumerate}[itemsep=0pt, topsep=0pt, leftmargin=12pt, labelsep=3pt, label=\arabic*.]
\item The neurons of the network are organized into distinct layers \m{\ell=0,1,\ldots L}.  
We will use \m{\actn{\ell}} to collectively denote the activations of all the neurons in layer \m{\ell}.  
\item The output of layer \m{\ell} can be expressed as 
\[\actn{\ell}=\xi_\ell(\phi_\ell(\actn{\ell-1})),\]
where \m{\phi_\ell} is a learnable \emph{linear} function, while \m{\xi_\ell} is a fixed nonlinearity. 
\item 
The entire network is equivariant to the action of a global symmetry group \m{G} in the sense that 
each \m{f_\ell} has a corresponding \m{G}--action that is equivariant to the action of \m{G} on the inputs. 
%equivariant to the action of global symmetry group \m{G} in the sense of Definition 
%\rf{eq: equiariance}. 
%of Definition 
%that if 
%\[x\mapsto T^{\textrm{in}}_g(x) \qquad \textrm{then}\qquad \actn{\ell}\mapsto T^\ell_g(\actn{\ell})
%\qquad\forall g\tin G
%\]
%for the appropriate \m{G}--actions \m{\cbrN{T^{\textrm{in}}_g}} and \m{\cbrN{T^\ell_g}}.  
\end{enumerate}
%\end{compactenum}
%Variations on equivariant networks of this type 
Generalized convolutional networks have found applications in a wide range of 
domains and their properties have been thoroughly investigated \citep{Cohen2016} 
\citep{Poczos2017} %\citep{KondorTrivedi2018} 
\citep{CohenCNNhomo}. 

\citet{KondorTrivedi2018} proved that in 
any neural network that follows the above axioms,  
the linear operation \m{\phi_\ell} must be a generalized form of covolution, 
as long as \m{\actn{\ell-1}} and \m{\actn{\ell}} can be conceived of as functions on 
quotient spaces \m{G/H_{\ell-1}} and \m{G/H_{\ell}} with associated actions  
%on which \m{G} acts the natural way:
\[\actf{}\overset{g}{\longmapsto} \actf{}^{g}\qqquad \actf{}^{g}(u)=\actf{}(g^{-1}u).\]%\qqquad g\tin G.\] 
%
%The following propositions show how activations transforming according to the first, 
%second and higher order permutation actions of \m{\Sn} can be mapped to such functions, 
%allowing us to take advantage of the this result. 
% \rf{eq: first order}
%and second order \rf{eq: second order} permutation actions can be regarded as functions on   
%that this is indeed the case for the first and second order permutation 
%actions, and that the corresponding quotient spaces 
%\m{\Sn/\Snm} and \m{\Sn/\Snmm}, respectively. 
%Recall that \m{(i,j)\tin\Sn} denotes the permutation that swaps \m{i} with \m{j} but leaves all other 
%elements of \m{\cbrN{\oneton{n}}} fixed. 
%
In this section we show that this result can be used to derive the most general form 
of convolutional networks whose activations transform according to \m{k}'th order permutation actions. 
Our key tools are 
Propositions 1--3 %\ref{prop: first order map}--\ref{prop: kth order map} 
in the Appendix, which show 
that the quotient spaces corresponding to the first, second and \m{k}'th order permutation 
actions are \m{\Sn/\Snm}, \m{\Sn/\Snmm} and \m{\Sn/\Sbb_{n-k}}:
\begin{enumerate}[itemsep=0pt, topsep=0pt, leftmargin=12pt, labelsep=3pt, label=\arabic*.]
\item \textbf{First order action.} If \m{f\tin\RR^d} is a vector 
transforming as %according to %on which \m{\Sn} acts by 
\m{f\overset{\sigma}{\longmapsto} f^\sigma} with 
\m{f^\sigma=\brN{f_{\sigma^{-1}(1)},\ldots,f_{\sigma^{-1}(n)}}}, 
then the corresponding quotient space function is 
\[\mathfrak{f}(\mu)=f_{\mu(n)} \qqquad \qqquad \mu\tin \Sn/\Snm.\]
\item \textbf{Second order action.}
If \m{f\tin\RR^{n\times n}} is a matrix with zero diagonal transforming as %on which \m{\Sn} acts by 
\m{f\overset{\sigma}{\longmapsto} f^\sigma} where 
\m{f^\sigma_{i,j}=f_{\sigma^{-1}(i),\ts \sigma^{-1}(j)}},\:\mbox{} 
then the corresponding quotient space function is 
\begin{equation*}%\label{eq: second order embedding}
\mathfrak{f}(\mu)=f_{\mu(n),\,\mu(n-1)} \qqquad \mu\tin\Sn/\Snmm.
\end{equation*}
\item \textbf{k'th order action.}
Let \m{f\tin\RR^{n\times \ldots\times n}} be a \m{k}'th order tensor %that is symmetric in all its indices 
such that \m{f_{\sseq{i}{k}}=0} unless \m{\sseq{i}{k}} are all distinct. 
If \m{f} transforms under permutations as %on which \m{\Sn} acts by 
\sm{f\overset{\sigma}{\longmapsto} f^\sigma} where 
\sm{f^\sigma_{\sseq{i}{k}}=f_{\sigma^{-1}(i_1),\ldots,\sigma^{-1}(i_k)}}, 
then the corresponding quotient space function is 
\begin{equation*}%\label{eq: kth order embedding}
\mathfrak{f}(\mu)=f_{\mu(n),\ldots,\mu(n-k+1)} \qqquad \mu\tin\Sn/\Sbb_{n-k}.
\end{equation*}
\end{enumerate}
%\bigskip
Theorem 1 of \citep{KondorTrivedi2018} states that 
%\citet{KondorTrivedi2018} prove that 
if the equivariant activations \m{\actn{\ell-1}} and \m{\actn{\ell}} 
both correspond to the same quotient space \m{S}, then the linear operation \m{\phi_\ell} must be of the form 
\[\phi_\ell(\mathfrak{f}_{\ell-1})=\mathfrak{f}_{\ell-1}\ast h_\ell,\]
where \m{\ast} denotes convolution on \m{\Sn}, and \m{h_\ell} is a function \m{S\backslash \Sn/S\to\RR}. 
Mapping \m{\phi(\mathfrak{f}_{\ell-1})} back to the original vector/matrix/tensor domain we get the 
following results. 
\ifdraft{\bigskip}

\begin{theorem}\label{thm: convo1}
If the \m{\ell}'th layer of a convolutional neural network maps a first order \m{\Sn}--equivariant 
vector \m{\fin\tin\RR^n} to a first order \m{\Sn}--equivariant vector \m{\fout\tin\RR^n}, 
then the functional form of the layer must be 
\[\ifdraft{f^\textrm{in}\mapsto f^{\textrm{out}}\qqquad} 
\fout_{i}=\xi\brbig{w_0 \fin_i+w_1 \fin_\ast},\]
where \sm{\fin_\ast\<=\sum_{k=1}^n \fin_k} and \m{w_0,w_1\tin\RR} are learnable weights. 
\end{theorem}
\smallskip 

\begin{theorem}\label{thm: convo2}
If the \m{\ell}'th layer of a convolutional neural network maps a second order \m{\Sn}--equivariant 
activation \m{\fin\tin\RR^{n\times n}} (with zero diagonal) to a second order \m{\Sn}--equivariant activation 
\m{\fout\tin\RR^{n\times n}} (with zero diagonal), then the functional form of the layer must be 
%\ificml{\m{\fin\mapsto \fout}}
\begin{multline*}
\ifdraft{
\fin\mapsto \fout\qqquad 
\fout_{i,j}=\xi\brbig{w_0 \fin_{i,j}+w_1 \fin_{j,i}+w_2 \fin_{i,\ast} +w_3 \fin_{\ast,i}+
w_4 \fin_{\ast,j}+w_5 \fin_{j,\ast}+w_6\fin_{\ast,\ast}}. 
}
\ificml{
\fout_{i,j}=\xi\brbig{w_0 \fin_{i,j}+w_1 \fin_{j,i}+w_2 \fin_{i,\ast} +w_3 \fin_{\ast,i}+\\
w_4 \fin_{\ast,j}+w_5 \fin_{j,\ast}+w_6\fin_{\ast,\ast}}, 
}
\end{multline*}
%\begin{multline*}
%\fout_{i,j}=\xi\brbig{w_0 \fin_{i,j}+w_1 \fin_{j,i}+w_2 \fin_{i,\ast} +w_3 \fin_{\ast,i}+\\
%w_4 \fin_{\ast,j}+w_5 \fin_{j,\ast}+w_6\fin_{\ast,\ast}}. 
%\end{multline*}
where \m{\fin_{p,\ast}\nts\<=\sum_k \fin_{p,k}}\ts,~\mbox{} \m{\fin_{\ast,p}\nts\<=\sum_k \fin_{k,p}}\ts,~\mbox{} 
\m{\fin_{\ast,\ast}\nts\<=\sum_{k,l} \fin_{k,l}},~\mbox{} and \m{w_0,\ldots,w_6\tin\RR} are learnable weights.
\end{theorem}

Theorem \ref{thm: convo1} is a restatement of the main result of \citep{DeepSets},  
which here we prove without relying on the Kolmogorov--Arnold theorem. 
Theorem \ref{thm: convo2} is its generalization to second order permutation actions. 
The form of third and higher order equivariant layers can be derived similarly, 
but the corresponding formulae are more complicated. These results also generalize naturally to 
the case when \m{\fin} and \m{\fout} have multiple channels. % (\promise{see Appendix}). 

Having to map each activation to a homogeneous space does put some restrictions on its form.  
For example, Theorem \ref{thm: convo2} requires that \m{f} have 
zeros on its diagonal, because the diagonal and off-diagonal parts of \m{f} actually form two 
separate homogeneous spaces. 
The Fourier formalism of the next section exposes the general case and removes these limitations.

\section{Fourier space activations}\label{sec: fourier}

Given any \m{\Sn}--action \sm{\setofN{T_\sigma}{\sigma\tin\Sn}}, for any \m{\sigma_1,\sigma_2}, we 
must have that \m{T_{\sigma_2\sigma_1}=T_{\sigma_2}T_{\sigma_1}}. 
This implies that 
$\cbrN{T_\sigma}_\sigma$ 
is a \emph{representation} of the symmetric group, 
%and allows us to 
bringing the full power of representation theory to bear on our problem 
%from representation to bear 
\citep{Serre}\citep{FultonHarris}. 
In particular, representation theory tells us that there is a unitary transformation \m{U} that 
simultaneously block diagonalizes all \m{T_\sigma} operators:  
\begin{equation}\label{decomp1}
U\ts T_\sigma\ts U^\dag=\bigoplus_{\lambda\vdash n}\bigoplus_{i=1}^{\kappa(\lambda)}\rho_\lambda(\sigma),
\end{equation}
%where \m{\cbrN{\rho_\lambda}_\lambda} 
where the \m{\rho_\lambda(\sigma)} matrix valued functions 
are the so-called irreducible representations (irreps) of \m{\Sn}. 
Here \m{\lambda\vdash n} means that \m{\lambda} ranges over the so-called \emph{integer partitions} of \m{n}, 
i.e., non-decreasing sequences of positive integers \sm{\lambda=(\sseq{\lambda}{k})} such that 
\sm{\sum_{i=1}^k\lambda_i\<=n}. It is convenient to depict integer partitions with so-called 
Young diagrams, such as 
\[\yng(4,3,1)\]
for $\lambda=(4,3,1)$.
It is a peculiar facet of the symmetric group that its irreps are best indexed by these combinatorial objects 
\citep{Sagan}. 
The integer $\kappa(\lambda)$ %called the multiplicity of \m{\rho_\lambda}, 
tells us how many times \m{\rho_\lambda} appears in the decomposition of the given action. 
Finally, we note that while in general the irreps of finite groups are complex valued, 
in the special case of the symmetric group they can be chosen to be real, allowing us to 
formulate everything in terms of real numbers.  

It is advantageous to put activations in the same basis that block diagonalizes the group action, 
i.e., to use \emph{Fourier space activations} \sm{\h f\<=U f}. 
Similar Fourier space ideas have proved crucial in the context of other group 
equivariant architectures
\cite{CohenSphericalICLR2018}\cite{SphericalCNNNeurIPS2018}. 
Numbering integer partitions and hence irreps according to inverse lexicographic order 
\m{(n)\<<(n\<-1,1)\<<(n\<-2,2)\<<(n\<-2,1,1)\<<\ldots} we define the \emph{type} 
\m{\tau=(\tau_1,\tau_2,\ldots)} of a given activation as the vector specifying how many times each 
irrep appears in the decomposition of the corresponding action. 
It is easy to see that if \m{f} is a first order permutation equivariant activation, 
then its type is \m{(1,1)}, whereas if it is a second order equivariant activation (with zero diagonal) 
then its type is \m{(1,2,1,1)}. The Fourier formalism allows further refinements. 
For example, if \m{f} is second order and symmetric, then its type reduces to \m{(1,2,1)}. 
On the other hand, if it has a non-zero diagonal, then its type will be \m{(2,3,1,1)}. 

It is natural to collect all parts of \sm{\h f} that correspond to the same irrep \m{\rho_{\lambda_i}} 
into a \m{d_{\lambda_i}\!\<\times \tau_i} matrix \m{F_{\lambda_i}}, 
where \m{d_{\lambda_i}} is dimension of \sm{\rho_\lambda} 
(i.e., \sm{\rho_\lambda(\sigma)\tin\RR^{d_\lambda\times d_\lambda}}).  
The real power of the Fourier formalism manifests in the following theorem, which gives 
a complete characterization of permutation equivariant convolutional architectures. 

\begin{theorem}\label{thm: fourier}
Assume that the input to a given layer of a convolutional type \m{\Sn}--equivariant network 
is of type \m{\tau=(\tau_1,\tau_2,\ldots)}, and the output is of type \m{\tau'=(\tau'_1,\tau'_2,\ldots)}. 
%Expressing the inputs and outputs of the linear part of the  
Then \m{\phi}, the linear part of the layer, in Fourier space must be of the form  
\[F_i\mapsto F_i W_i,\]
where \sm{\cbrN{F_i\tin \RR^{d_{\lambda_i}\times \tau_i}}_i} are the Fourier matrices of the input to the layer, 
and \sm{\cbrN{W_i\tin \RR^{\tau_i\times \tau'_i}}_i} are (learnable) weight matrices. 
In particular, the total number of learnable parameters is \m{\sum_i \tau_i\times \tau'_i}.
\end{theorem}

This theorem is similar to the results of \citep{KondorTrivedi2018}, but is more general 
because it does not restrict the actiations to be   
%it allows for the case when the activations are not just 
functions on individual homogeneous spaces. 
In particular, it makes it easy to count the number of allowable parameters in any layer, and 
derive the form of the layer, such as in the following corollary. 

\begin{corollary}\label{coro: second order}
If a given layer of a convolutional neural network maps a second order \m{\Sn}--equivariant 
activation \m{\fin\tin\RR^{n\times n}} \ignore{f_{\ell-1}\tin\RR^n}(with non-zero diagonal) 
to a second order \m{\Sn}--equivariant activation \m{\fout\tin\RR^{n\times n}} 
\ignore{f_\ell\tin\RR^n}(with non-zero diagonal), 
then the \ifdraft{functional form of the }layer's operation must be 
%\sm{\fin\<\mapsto \fout} with 
\begin{multline*}
\ifdraft{
\fout_{i,j}=\xi\brbig{w_0 \fin_{i,j}+w_1 \fin_{j,i}+w_2 \fin_{i,\ast} +w_3 \fin_{\ast,i}+
w_4 \fin_{\ast,j}+w_5 \fin_{j,\ast}+w_6\fin_{\ast,\ast}+\\ 
w_7 \wbar{\fin_{\ast}}+w_8\fin_{i,i}+w_9\fin_{j,j}+
\delta_{i,j}\brN{w_{10} \wbar{\fin_{i}}+ w_{11}\wbar{\fin_{\ast}}+w_{12}\fin_{\ast,\ast}
+w_{13}\fin_{i,\ast}+w_{14}\fin_{\ast,i}}}. 
}
\ificml{
\fout_{i,j}=\xi\brbig{w_0 \fin_{i,j}+w_1 \fin_{j,i}+w_2 \fin_{i,\ast} +w_3 \fin_{\ast,i}+w_4 \fin_{\ast,j}+\\
w_5 \fin_{j,\ast}+w_6\fin_{\ast,\ast}+w_7 \wbar{\fin_{\ast}}+w_8\fin_{i,i}+w_9\fin_{j,j}+\\
\delta_{i,j}\brN{w_{10} \wbar{\fin_{i}}+ w_{11}\wbar{\fin_{\ast}}+w_{12}\fin_{\ast,\ast}
+w_{13}\fin_{i,\ast}+w_{14}\fin_{\ast,i}}}. 
}
\end{multline*}
where \m{\fin_{p,\ast}\!\<=\!\sum_k \fin_{p,k}}\ts,\mbox{} \m{\fin_{\ast,p}\!\<=\!\sum_k \fin_{k,p}}\ts,\mbox{} 
\m{\fin_{\ast,\ast}\!\<=\!\sum_{k,l} \fin_{k,l}}\ts,\mbox{}
\m{\wbar{\fin_p}\!\<=\fin_{p,p}},\mbox{} and \sm{\wbar{\fin_\ast}\<=\sum_k \fin_{k,k}}.  
Here \m{w_0,\ldots,w_{14}} are learnable parameters.  
\end{corollary}

This case was also discussed in \cite{maron2018invariant}, Appendix A.
In the general case, computing the \m{\sseq{F}{p}} matrices from \sm{\fin} requires a (partial) \m{\Sn}--Fourier 
transform, 
whereas computing the \sm{\fout} from the \m{F_i W_i} matrices requires an inverse Fourier transform. 
Fast \m{\Sn} Fourier transforms have been developed that can accomplish these transformations in 
\m{O(k s n^2)} operations, where \m{k} is the order of the action and \m{s=\sum_{i}\tau_i} 
is the total size of the input (or output) activation \citep{Clausen89}\citep{MaslenRockmoreSeparationI}. 
Unfortunately this topic is beyond the scope of the present paper.   
We note that since the Fourier transform is a unitary transformation, the notion of \m{\Sn}--\emph{type} of 
an activation and the statement in Theorem \ref{thm: fourier} regarding the number of learnable 
parameters remains valid no matter whether a given neural network actually stores activations 
in Fourier form.

\section{Compositional networks}\label{sec: compositional} 

%\citep{DeepSets}, XXX and XXX 
Most prior work on \m{\Sn}--equivariant networks 
considers the case where each activation is acted on by the entire group \m{\Sn}. 
In many cases, however, a given neuron's ouput only depends on a subset of objects, and therefore 
we should only consider a subgroup \m{\Sbb_k} of \m{\Sn}. 
Most notably, this is the case in message passing graph neural networks (MPNNs) \citep{Riley2017}, where the 
output of a given neuron in the \m{\ell}'th layer only depends on a neighborhood of the corresponding 
vertex in the graph of radius \m{\ell}. 
Of course, as stated in the Introduction, in most existing MPNNs 
the activations are simply invariant to permutations. However, there have been attempts in the literature 
to build covariance into graph neural networks, such as in \citep{CCN-JCP}, where,   
in the language of the present paper, the internal activations are first or second order permutation equivariant. 
Another natural example are networks that learn rankings by fusing partial rankings. % \risi{(cite)}. 
%We have the following general definition. 
%We will use the following definition taken from \citep{CCN-JCP}. 

\begin{definition}
Let \m{\Ocal=\cbrN{\sseq{o}{n}}} be a set of \m{n} objects that serve as inputs to a feed-forward neural 
network \sm{\Ncal}. 
Given a neuron \m{\neuron}, we define the domain \m{\domr(\neuron)} of \m{\neuron} 
as the largest ordered subset of \sm{\Ocal} such that  
%as the largest \sm{\Ocal_\neuron\subseteq \Ocal} such that  
the ouput of \m{\neuron} does not depend on \sm{\Ocal\setminus\Ocal_\neuron}. A 
\df{permutation covariant compositional neural network} is a neural network in which: 
\begin{compactenum}[~~1.]
\item The neurons form a partially ordered set such that if \sm{\neuron_{c_1},\ldots,\neuron_{c_k}} form 
the inputs to \m{\neuron}, then \m{\domr(\neuron)=\bigcup_{i}\domr(\neuron_{c_i})}. 
\item Under the action of \m{\sigma\tin \Sn} on \m{\Ocal}, 
%the output of each neuron \m{\neuron} 
%with \m{\absN{\domr{(\neuron)}}\<=k} 
%transforms according to a covariant action of \sm{\Sbb_{\abs{\domr(\neuron)}}}.  
%on \m{\domr(\neuron)\subsete} with \m{\sigma}.   
the network transforms to \m{\Ncal'}, such that for each \m{\neuron\tin\Ncal} there is a corresponding 
\m{\neuron'\tin\Ncal'} with \m{\domr(\neuron')=\pi(\sigma(\domr(\neuron)))} for some 
\m{\pi\tin\Sbb_{\abs{\domr{\neuron}}}}.
\item Each neuron has a corresponding action \sm{\setofN{T^\neuron_\pi}{\pi\tin\Sbb_{\abs{\domr(\neuron)}}}} 
such that \m{f_{\neuron'}=T^\neuron_\pi(f_{\neuron'})}. 
\end{compactenum}
\end{definition}
\begin{figure}
\includepicc{.55}{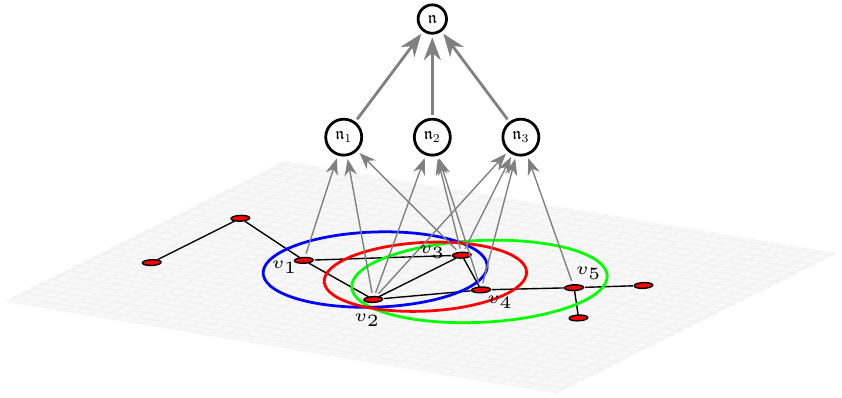}
\vspace{-12pt}
\caption{
In a compositional network each neuron \m{\neuron} is only 
equivariant to the permutations of a subset of objects \m{\domr(\neuron)\subseteq \Ocal}. 
Here, for example, \m{\neuron_1} captures information from vertices \m{\cbrN{v_1,v_2,v_3}}; \m{\neuron_2} 
from \m{\cbrN{v_2,v_3,v_4}}; and \m{\neuron_3} from \m{\cbrN{v_2,v_3,v_4,v_5}}. 
When the outputs of these three neurons are aggregated in a higher level neuron,  
whose domain covers %\m{\cbrN{v_1,v_2,v_3,v_4,v_5}}, 
all five vertices, 
the outputs of \m{\neuron_1,\neuron_2} and \m{\neuron_3} must be 
\emph{promoted} to transform equivariantly wrt.\;\m{\Sbb_5}. 
\label{fig: compositional}}
\vspace{-0.2in}
\end{figure}

The above definition captures the natural way in which any network  
that involves a hierarchical decomposition of a complex objects into parts is expected to behave. 
Informally, the definition says that each neuron is only responsible for a subset 
of objects (\sm{\domr(\neuron)}) and if the objects are permuted by \m{\sigma}, then 
the output of the neuron is going to transform according the restriction of \m{\sigma} to this subset, 
which is \m{\pi}.   
For example, message passing neural networks follow this schema with \m{\domr(\neuron)} being the 
\m{1}-neighborhoods, \m{2}-neighborhoods, etc., of each vertex. 
However, in most MPNNs the activations are \emph{invariant} under permutations, which corresponds to 
\m{\cbrN{T^\neuron_\pi}} being the trivial action of \sm{\Sbb_{\abs{\domr{\neuron}}}}. 
If we want to capture a richer set of interactions, specifically interactions where the output of \m{\neuron} 
retains information about the individual identities of the objects in its domain, relations between 
the objects, or their relative rankings, for example, then it is essential that the outputs of the neurons 
be allowed to vary with higher order permutation actions (Figure \ref{fig: compositional}).

Each neuron in a compositional architecture aggregates information related to 
\sm{\domr(\neuron_{c_1}),\ldots,\domr(\neuron_{c_k})\subseteq \domr(\neuron)} to 
\sm{\domr(\neuron)} itself. 
Being able to do this without losing equivariance requires knowing the correspondence between 
each element of \sm{\domr(\neuron_{c_p})} and \sm{\domr(\neuron)}. 
To be explicit, we will denote which element of \sm{\domr(\neuron)} any given \sm{i\tin \domr(\neuron_{c_p})} 
corresponds to by \m{\chi^\neuron_p(i)}, and we will call \emph{promotion} the act of reindexing each incoming 
activation \sm{\actn{\neuron_{c_p}}} to transform covariantly with permutations of 
the larger set \sm{\domr(\neuron)}. 
The promoted version of \sm{\actn{\neuron_{c_p}}} we denote \sm{\widetilde{\actn{}}{}^{\neuron_{c_p}}}. 
In the case of first, second, etc., order activations stored in vector/matrix/tensor form, promotion 
corresponds to simple reindexing:
\begin{eqnarray}\label{eq: promotion}
%\widetilde{\actn{}}{}^{\neuron_{c_p}}_i=\actn{\neuron_{c_p}}_{(\chi^\neuron_p)^{-1}(i)}
%\qqquad 
%\widetilde{\actn{}}{}^{\neuron_{c_p}}_{i,j}=\actn{\neuron_{c_p}}_{(\chi^\neuron_p)^{-1}(i),\ts(\chi^\neuron_p)^{-1}(j)}
[\widetilde{\actn{}}{}^{\neuron_{c_p}}]_{\chi^\neuron_p(i)}=[\actn{\neuron_{c_p}}]_{i},
\quad \ifdraft{\qquad\qquad}
[\widetilde{\actn{}}{}^{\neuron_{c_p}}]_{\chi^\neuron_p(i),\ts\chi^\neuron_p(j)}=[\actn{\neuron_{c_p}}]_{i,j}
\ifdraft{, \qquad \textrm{etc.}}.
\end{eqnarray}
In the case of Fourier space activations, with the help of FFT methods, computing each \sm{\widetilde{F}{}^{\neuron_{c_p}}_{i}}  
takes \sm{O((\absN{\nts\domr(\neuron)}-\absN{\nts\domr(\neuron_{c_p})})\ts d_{\lambda_i}\ts \tau_i)} operations. 
Describing this process in detail is unforutnately beyond the scope of the present paper. 

Once promoted, \sm{\widetilde{\actn{}}{}^{\neuron_{c_1}},\ldots,\widetilde{\actn{}}{}^{\neuron_{c_k}}} 
must be combined in a way that is itself covariant with permutations of \m{\domr(\neuron)}. 
The simple way to do this is to just to sum them. For example, in a second order network, using 
Theorem \ref{thm: convo2} we might have 
\ifdraft{
\[
\fout_{i,j}=\xi\brbig{w_0 \finb_{i,j}+w_1 \finb_{j,i}+w_2 \finb_{i,\ast} +w_3 \finb_{\ast,i}+
w_4 \finb_{\ast,j}+w_5 \finb_{j,\ast}+w_6\finb_{\ast,\ast}},  
\]
}
\ificml{
\begin{multline*}
\fout_{i,j}=\xi\brbig{w_0 \finb_{i,j}+w_1 \finb_{j,i}+w_2 \finb_{i,\ast} +w_3 \finb_{\ast,i}+\\
w_4 \finb_{\ast,j}+w_5 \finb_{j,\ast}+w_6\finb_{\ast,\ast}},  
\end{multline*}
}
where \m{\finb=\sum_{i}\widetilde{\actn{}}{}^{\neuron_{c_k}}}. 
This type of aggregation rule is similar to the summation operation used in MPNNs. 
In particular, while it correctly accounts for the overlap between the domains of upstream 
neurons, it does not take into account which \sm{\widetilde{\actn{}}{}^{\neuron_{c_i}}} 
relates to which subset of \m{\domr(\neuron)}. 
A richer class of models is obtained by forming potentially higher order covariant \emph{products} of the 
promoted incoming activations rather than just sums. 
Such networks are still covariant but go beyond the mold of convolutional nets because by virtue of the 
nonlinear interaction, one input to the given neuron can \emph{modulate} the other inputs. 
% \promise{One example of this appears in the next section.}  

\section{Second order permutation equivariant variational graph auto-encoder}\label{sec: vae}

% Ultimately, it is the combination of higher order permutation equivariance with the compositional 
% architecture described in the previous section that gives rise to expressive networks. 
To demonstrate the power of higher-order permutation-equivariant layers,
we apply them to the problem of graph learning, where the meaning of both 
equivariance and compositionality is very intuitive. 
We focus on generative models rather than just learning from graphs, 
as the exact structure of the graph is uncertain.
Consequently, edges over all possible pairs of nodes must, at least in theory, be considered.
This makes a natural connection with the permutation group.
% because in the decoding phase, where larger structures are generated by stitching together 
%partially overlapping smaller structures, equivariance is especially important. 
%we propose an equivariant variational encoder, 
%because, as we will see, in the decoding phase equivariance is especially important. 

Variational Auto-encoders (VAE's) 
%learn a low dimensional representation of graphs 
%in a low dimensional space and a distribution over that space 
consist of an encoder/decoder pair where the encoder learns a low dimensional representation of, 
while the decoder learns to reconstruct using a simple probabilistic model on a latent representation.\cite{kingma2013auto}
The objective function combines two terms: the first term measures how well 
the decoder can reconstruct each input in the training set from its latent representation, whereas 
%each graph in the training set can be reconstructed from its latent representation by the decoder, whereas 
the second term aims to ensure that the distribution corresponding to the training data is as close 
as possible to some fixed target distribution such as a multivariate normal. After training, 
sampling from the target distribution, possibly with constraints, will generate samples 
% from realistic graphs 
that are similar to those in the training set. 
% VGAE's are commonly used for e.g., generating organic molecules with specific properties. 

Equivariance is important when constructing VAE's for graphs because in order to compare each input graph to the output 
graph generated by the decoder, the network needs to know which input vertex corresponds to which output vertex.
This allows the quality of the reconstruction to be measured without solving an expensive graph-matching problem.
The first architectures were based on spectral ideas, in which case equivariance comes 
``for free'' \citep{kipf2016variational}. 
%In VGAE's   based on spectral ideas such as \cite{kipf2016variational}, this property comes ``for free'', 
%since each spectral convolutional layer is naturally equivariant. 
However, spectral VGAE's are limited in the extent to which they can capture the combinatorial 
structure of graphs because they draw every edge independently 
(albeit with different parameters). 

An alternative approach is to generate the output graph in a sequential manner, somewhat similarly to 
how RNNs and LSTMs generate word sequences 
% \citep{JunctionTreeVAE_ICML2018}
\citep{GCPN_NIPS2018}\citep{GRAN_NIPS2019}. 
These methods can incorporate rich local information in the generative process, 
such as a library of known functional groups in chemistry applications. 
On the other hand, this can break the direct correspondence between the vertices of the input graph and 
the output graph, so the loss function can only compare them via global graph properties
%case there is no direct relationship between the vertices of the input graph and the output graph, so 
%instead they have to be compared by some graph properties 
such as the frequency of certain small subgraphs, etc..  
%Our aim is to construct a VGAE that better captures the combinatorial 
%Using the \m{\Sn}--equivariance formalism 
% Our aim in this section we construct a VGAE that similarly to these methods has a local combinatorial flavor, 
% yet is still end-to-end equivariant.

\subsection{The latent layer}

To ensure end-to-end equivariance, we set the latent layer to be a matrix  \m{P\tin\RR^{n\times C}}, 
whose first index is first order equivariant with permutations. At training time, \m{P} is computed from 
\sm{\actn{\neuron^{\textrm{top}}}_{i,j,a}} simply by summing over the \m{j} index. 
For generating new graphs, however, we must specify a distribution for \m{P} that is 
\emph{invariant} to permutations. In the language of probability theory, such distributions are 
called \emph{exchangeable} \cite{kallenberg2006probabilistic}.  

For infinite exchangeable sequences de Finetti's theorem states that any such distribution is a mixture of 
iid.~distributions \cite{deFinetti1930}. 
For finite dimensional sequences the situation is somewhat more complicated \cite{diaconis1980finite}. 
Recently a number of researchers have revisited exchangeability from the perspective of machine learning, 
and specifically graph generation \citep{OrbanzRoy2015}\citep{CaiNIPS2016}\citep{BloemReddy2019}.

%\section{Experiments}
\label{sec: experiments}

%\subsection{Link prediction on citation graphs}
\label{sec:link_prediction}

\paragraph{Link prediction on citation graphs.} 
% We demonstrate the ability of the SnNN models to learn meaningful latent embeddings on a link prediction task on popular citation network datasets Cora and Citeseer \citep{PSen2008}. In the training time, 15\% of the citation links (edges) have been removed while all node features are kept, the models are trained on an incomplete graph Laplacian constructed from the rest 85\% of the edges. From previously removed edges, we sample the same number of pairs of unconnected nodes (non-edges). We form the validation and test sets that contain 5\% and 10\% of edges with an equal amount of non-edges, respectively. Hyperparameters optimization (e.g. number of layers, dimension of the latent representation, etc.) is done on the validation set. \\ \\
% We compare our model SnNN against spectral clustering (SC) \citep{Tang2011}, deep walk (DW) \citep{Perozzi2014}, non-probabilistic graph autoencoder (GAE) and variational graph autoencoder (VGAE) \citep{kipf2016variational} on the ability to correctly classify edges and non-edges using two metrics: area under the ROC curve (AUC) and average precision (AP). Numerical results of SC, DW, GAE and VGAE and experimental settings are taken from \citep{kipf2016variational}. For SnNN, we initialize weights by Glorot initialization \citepGlorot2010}. We train for 2,048 epochs using Adam optimization \citepKingma2015} with a starting learning rate of $0.01$. The number of layers range from 1 to 4. The size of latent representation is 64. \\ \\
To demonstrate the efficacy of higher order \m{\Sn}-equivariant layers, we compose them with the original VGAE architecture in \citep{kipf2016variational}.
While this architecture is a key historical benchmark on link-prediction tasks, 
it has the drawback that the encoder takes the form $\xi\left( z z^\top \right)$, where $z$ is the value of the graph encoded in the latent layer.
Effectively, the decoder of the graph has no hidden layers, which limits its expressive power.
However, we show that if this can be mitigated by composing the same architecture with \m{\Sn}-convolutional layers.
Specifically, we place additional convolutional layers between the outer product $z z^\top$ and the final sigmoid, interspersed with ReLU nonlinearities.
We then apply the resulting network to link prediction on the citation network datasets Cora and Citeseer \citep{PSen2008}. In training time, 15\% of the citation links (edges) have been removed while all node features are kept.
The models are trained on an incomplete graph Laplacian constructed from the remaining 85\% of the edges. 
From previously removed edges, we sample the same number of pairs of unconnected nodes (non-edges). 
We form the validation and test sets that contain 5\% and 10\% of edges with an equal amount of non-edges, respectively. 
Hyperparameters optimization (e.g. number of layers, dimension of the latent representation, etc.) is done on the validation set. 

We then compare the architecture against the original (variational) graph autoencoder \citep{kipf2016variational}, as well as spectral clustering (SC) \citep{Tang2011}, deep walk (DW) \citep{Perozzi2014}, and GraphStar \citep{haonan2019graph} and compare the ability to correctly classify edges and non-edges using two metrics: area under the ROC curve (AUC) and average precision (AP). Numerical results of SC, DW, GAE and VGAE and experimental settings are taken from \citep{kipf2016variational}. We initialize weights by Glorot initialization \cite{Glorot2010}. We train for 2,048 epochs using Adam optimization \cite{Kingma2015} with a starting learning rate of $0.01$. The number of layers range from 1 to 4. The size of latent representation is 64. 

\begin{table}[t]
\begin{center}
{\small 
\begin{tabular}{||l | c | c ||}
	\hline
	Method & AUC & AP \\
	\hline\hline
	SC & 84.6 $\pm$ 0.01 & 88.5 $\pm$ 0.00 \\
	\hline
	DW & 83.1 $\pm$ 0.01 & 85.0 $\pm$ 0.00 \\
	\hline
	GAE & 91.0 $\pm$ 0.02 & 92.0 $\pm$ 0.03 \\
	\hline
	VGAE & 91.4 $\pm$ 0.01 & 92.6 $\pm$ 0.01 \\
	\hline
    GraphStar & {95.65} $\pm$ ? & {96.15} $\pm$ ? \\
	\hline
%    VGAE + 1 SnConv & 94.6 $\pm$ 0.02 & 93.9 $\pm$ 0.02 \\
%    \hline 
    % VGAE + 4 SnConv (best n) & 94.6 $\pm$ 0.02 & 93.9 $\pm$ 0.02 \\
    2nd order VGAE (our method) & \textbf{96.1} $\pm$ 0.07 & \textbf{96.4} $\pm$ 0.06 \\
	\hline
\end{tabular}
}
\end{center}
\vspace{-8pt}
\caption{\label{tbl:cora} Cora link prediction results (AUC \& AP)}
\vspace{-12pt}
\end{table}

\begin{table}[t]
\begin{center}
{\small 
\begin{tabular}{||l | c | c ||}
	\hline
	Method & AUC & AP \\
	\hline\hline
	SC & 80.5 $\pm$ 0.01 & 85.0 $\pm$ 0.01 \\
	\hline
	DW & 80.5 $\pm$ 0.02 & 83.6 $\pm$ 0.01 \\
	\hline
	GAE & 89.5 $\pm$ 0.04 & 89.9 $\pm$ 0.05 \\
	\hline
	VGAE & 90.8 $\pm$ 0.02 & 92.0 $\pm$ 0.02 \\
	\hline
    GraphStar & \textbf{97.47} $\pm$ ? & \textbf{97.93} $\pm$  ? \\
	\hline
%    VGAE + 1 SnConv& 92.7 $\pm$ 0.04 & 92.4 $\pm$ 0.03 \\
%    \hline
    2nd order VGAE (our method)& 95.3 $\pm$ 0.02 & 94.3 $\pm$ 0.02 \\
	\hline
\end{tabular}
}
\end{center}
\vspace{-8pt}
\caption{\label{tbl:citeseer} Citeseer link prediction results (AUC \& AP)}
\vspace{-12pt}
\end{table}

We found that these additional layers make VGAE so expressive it quickly overfits.  We consequently regularize the models through early stopping.
Tables \ref{tbl:cora} and \ref{tbl:citeseer} show our numerical results in Cora and Citeseer datasets.  We give the performance of our best model for each task, as well as the performance from only adding a single layer to VGAE.
% SnNN outperforms all other baselines, especially VGAE with a considerable margin of 5\% in AUC and 2\% in AP.
Our results improve over the original VGAE architecture in all categories. For Cora, VGAE becomes competitive with more recent architectures such as GraphStar.
Moreover, we see that already a single layer of the second-order equivariant gives a tangible improvement over the previous VGAE results.

%\subsection{Molecular generation}
\label{sec:molecular-generation}

\paragraph{Molecular generation.}
% We compare SnNN against ... in the task of molecular generation. 
We next explore the ability of $\Sn$-equivariant layers to build graphs with 
highly structured graphs
% considerable local structure
by applying them to the task of molecular generation.
% Indeed, we note in contrast to Message-Passing architectures, the layers in \ref{coro: first order} only involve comparisons over the same node
% A key point of our work is that we incorporate no domain knowledge in our architecture:
We do not incorporate any domain knowledge in our architecture; the only input to the network is a one-hot vector of the atom identities and the adjacency matrix.
% any information about molecular structure is learned directly by the network.
As we intend to only test the expressive power of the convolutional layers, here we construct both the encoder and the decoder 
using activations of the form in~\ref{coro: second order} with ReLU nonlinearities.
Node features can be incorporated naturally as a second-order features with zero off-diagonal elements.
Similar to (V)GAE, we use a latent layer of shape $n \times c$, where $c$ is a the size of the latent embedding for each node.
To do form the values in the latent layer, we take a linear combination of first-order terms in Corollary~\ref{coro: second order},
\begin{multline}\label{eq:two_to_one}
    f^{\textrm{latent}}_{i}=\xi\brbig{w_1 \fin_{i,\ast} +w_2 \fin_{\ast,i}+w_3\fin_{\ast,\ast}+w_4 \wbar{\fin_{\ast}}+w_5\fin_{i,i}}. 
\end{multline}
We take our target distribution in the variational autoencoder to be \sm{\mathcal{N}\left(0,1\right)} on this space, independently on all of the entries.
To reconstruct a second-order feature from the latent layer in the decoder, we take the tensor product of each channel of $f^{\textrm{latent}}_{i}$ to build a collection of rank one second-order features.
The rest of the decoder then consists of layers formed as described in Corollary~\ref{coro: second order}.
In the final layer we apply a sigmoid to a predicted tensor of size $n \times n \times 3$. 
This tensor represents our estimate of the labelled adjacency matrix, which should $1$ in the $i,j,k$'th element if atoms $i$ and $j$ are connected by a bond of type $k$ in the (kekulized) molecule.
Similarly, we also predict atom labels by constructing a first-order feature as in \eqref{eq:two_to_one} with the number of channels equal to the number of possible atoms and apply a softmax across channels.
The accuracy of our estimate is then evaluated using a binary cross-entropy loss to the predicted bonds and a negative log-likelihood loss to the predicted atom labels.
To actually generate the molecule, we choose the atom label with largest value in the softmax, and connect it with all other atoms where our bond prediction tensor is larger than $0.5$.

% demonstrates the ability of networks using higher-order equivariant layers to learn more complex structures.

% We build a variational autoencoder using the second-order equivariant activation described in section \ref{sec: vae}.
To test this architecture, we apply it to the ZINC dataset \citep{Sterling2015} and attempt to both encode the molecules in the dataset and generate new molecules.  
Both the encoder and the decoder have 4 layers of with 100 channel indices with a ReLU nonlinearity, and we set the latent space to be of dimension $n_{max} \times 20$, where $n_{max}$ is the largest molecule in the dataset (smaller molecules are treated using disconnected ghost atoms).
We train for 256 epochs, again using the Adam algorithm.\cite{Kingma2015}
Interestingly, we achieved better reconstruction accuracy and generated molecules when disabling the KL divergence term in the VAE loss function.
We discuss potential reasons for this in the supplement.

To evaluate the accuracy of the encodings, we evaluate the reconstruction accuracy using the testing split described in \citep{pmlr-v70-kusner17a}.   
% We then determine how many correspond to valid %   of molecules in the test set, and validity of new molecules generated from the prior, .
% Our prior consists of a \sm{\mathcal{N}\left(0,1\right)} distribution on a \sm{n \times 7} dimensional space, where $n$ is the number of nodes in the largest graph considered (see section \ref{sec:latent}).
% The final layer of our encoder mixes down to 14 channels, 7 of which correspond to the means and 7 of which correspond to the variances in the approximate posterior $q_\phi$ of the variational autoencoder.
% We choose our latent space to be of dimension 7.
% All models were trained and validated on ZINC dataset \citep{Sterling2015} using the testing split described in \citep{pmlr-v70-kusner17a}.
In table~\ref{tbl:zinc}, we report numerical results for two metrics: reconstruction accuracy of molecules in the test set, and validity of new molecules generated from the prior
and compare with other algorithms.
\begin{table}[t]
\begin{center}
\small{
\begin{tabular}{||l | c | c ||}
	\hline
	Method & Accuracy & Validity \\ 
	\hline\hline
	CVAE & 44.6\% & 0.7\% \\
	\hline
	GVAE & 53.7\% & 7.2\% \\
	\hline
	SD-VAE & 76.2\% & 43.5\% \\
	\hline
	GraphVAE & - & 13.5\% \\
    \hline
	Atom-by-Atom LSTM & - & 89.2\% \\
	\hline
    JT-VAE & 76.7 \% & \textbf{100.0}\% \\
	\hline
	% 2nd order $\Sn$-Conv VAE (our method) & {76.0\%} & {44\%} \\
    2nd order $\Sn$-Conv VAE (our method) & \textbf{98.4\%} & {33.4\%} \\
	\hline
\end{tabular}
}
\end{center}
\vspace{-8pt}
\caption{\label{tbl:zinc} ZINC results (reconstruction accuracy \& validity)}
\vspace{-18pt}
\end{table}

\ignore{
\begin{table}[t]
\caption{\label{tbl:qm9} QM9 results (reconstruction accuracy \& validity)}
\begin{tabular}{||l | c | c ||}
	\hline
	Method & Reconstruction accuracy & Validity \\
	\hline\hline
	\hline
\end{tabular}
\end{table}
}
% We find that our algorithm has similar reconstruction accuracy to our baselines of Junction-Tree variational autoencoder (JT-VAE) \citep{DBLP:journals/corr/abs-1802-04364} and Syntax-Directed variational autoencoder (SD-VAE) \citep{DBLP:journals/corr/abs-1802-08786}.
Our algorithm gives the best reconstruction among the accuracies considered.
Moreover, we note that our algorithm requires considerably less domain knowledge than many of the other algorithms in Table \ref{tbl:zinc}. 
% Both JT-VAE and the Atom-by-Atom LSTM \citep{DBLP:journals/corr/abs-1803-03324} construct more valid molecules.
The SD-VAE, Grammar VAE (GVAE) \citep{pmlr-v70-kusner17a}, and Character VAE (CVAE) \citep{Bombarelli2016} treat the task of molecular graph generation as text generation in Natural Language Processing by explicitly using the SMILES strings: structured representations of molecules that, in their construction, encode rules about chemical connectivity.
JT-VAE explicitly draws from a vocabulary of chemical structures and explicitly ensures that the output obeys known chemical rules.
In contrast, our model is given no knowledge of basic chemical principles such as valency or of functional groups, and must learn it from first principles.
% We also ran the model on a smaller latent layer of $n_{max} \times 7$ dimensions; this is gave a reconstruction accuracy of $76 \%$ and a validity of $44 \%$. However, many of the constructed molecules are comparatively small

% We next test the ability of the architecture to generate new molecules.  We sample 5000 IID normal vectors from the prior, and construct the corresponding graphs from the decoder.
To test the validity of newly generated molecules, we take 5000 samples from our target distribution, feed them into the decoder, and determine how many correspond to valid SMILES strings using RDKit.
Our model does not learn to construct construct a single, connected graph; consequently we take the largest connected component of the graph as our predicted molecule.
% We suspect this procedure, , this procedure biases our results towards small molecules: 
Perhaps a result of our procedure for reading the continuous bond features into a single molecule with discrete bonds, we find that our results are biased towards small molecules:
roughly 20 percent of the valid molecules we generate have 5 heavy atoms or less.
In figure~\ref{fig:large_molecules} we give a random sample of 12 of the generated molecules that are syntactically valid and have more than 10 atoms.
% Surprisingly, we see that our architecture learns to construct 6 atom rings, a sterically preferred cyclical arrangement.
% This occurs despite the fact that many-atom rings would naively seem to be fundamentally higher-order feature.
Comparing with the molecules in the ZINC dataset, our network constructs many molecules that are sterically strained. 
Many generated molecules have 4-atom rings, features which are rare seen in stable organic molecules.
Indeed, the ability to make inferences over types of molecular rings is an example of higher-order information, requiring direct comparison over multiple atoms at once.
However, we note that our model is able to construct molecules with relatively reasonable bonding patterns and carbon backbones.
In conjunction with our reconstruction results, we believe that even simple higher-order information can be used to infer more complex graph structure.

\begin{figure}
%\includepicc{.5}{../images/large_mols_img.pdf}
\includepicc{.5}{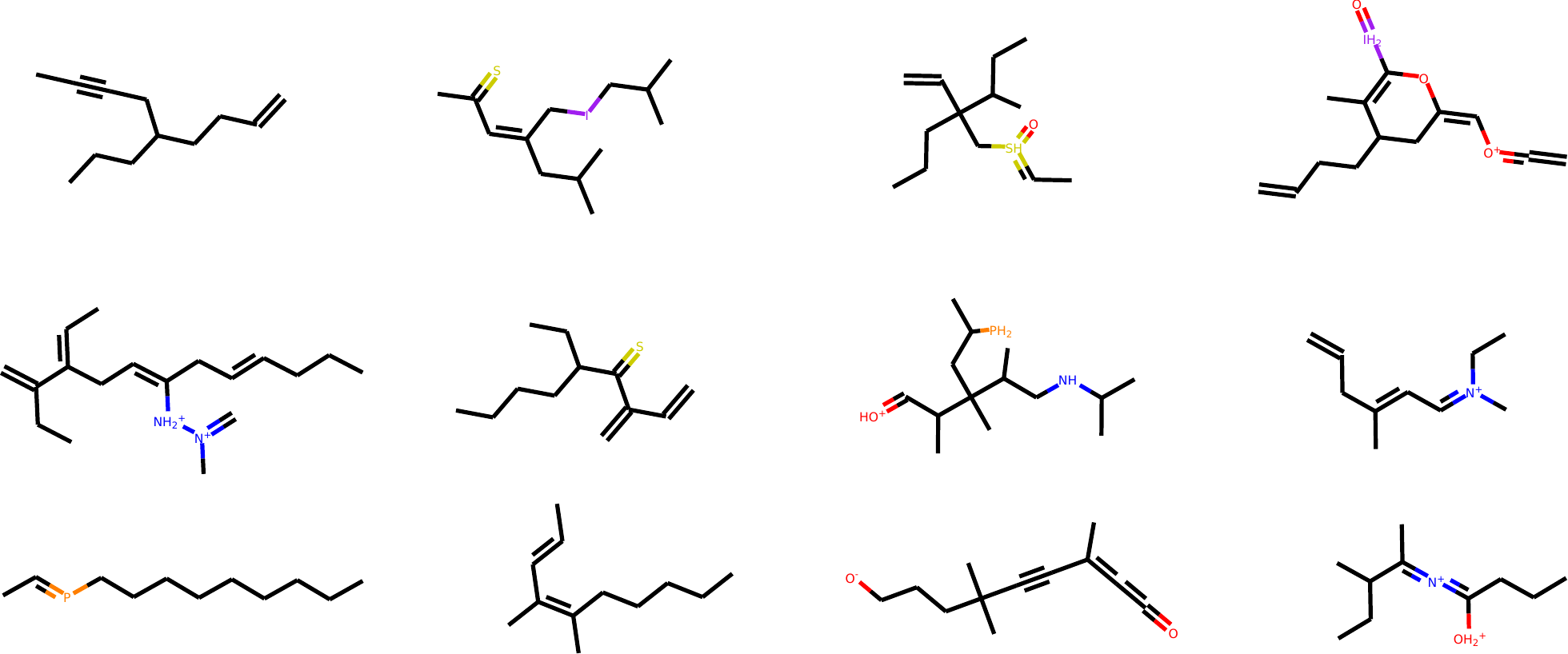}
\vspace{-2pt}
\caption{
A sample of 12 molecules of generated by the second-order graph variational autoencoder.
Molecules are constrained to be syntactically valid (can be converted into valid SMILES strings) and have more than 10 atoms.
}
%\label{fig:large_molecules}}
\label{fig:large_molecules}
\vspace{-0.2in}
\end{figure}

% Real molecules are more likely to have rings with 5-8 atoms.  
% These patterns are clear examples of higher-order features, and we theorize that future work that uses higher-order forms of equivariance will improve on these aspects of predictions.
% However, we also note that our predictions have clear carbon-backbones

%\input{link_prediction}
%\input{molecular_generation}
 
\section{Conclusions}

In this paper we argued that to fully take into account the action of permutations on neural networks, 
one must consider not just the defining representation of \m{\Sn}, but also some of the higher order irreps. 
This situation is similar to, for example, recent work on spherical CNNs, where images on the surface of the 
sphere are expressed in terms of spherical harmonics up to a specific order \cite{CohenSphericalICLR2018}. 
Specifically, to correctly learn \m{k}'th order relations between members of a set requires a \m{k}'th 
order permutation equivariant architecture. 

Graphs are but the simplest case of this, since whether or not two vertices are connected by an edge 
is a second order relationship. Already in this case, our experiments on link prediction and molecule 
generation show that equivariance gives us an edge over other algorithms, which is noteworthy 
given that the other algorithms use hand crafted chemical features (or rules). 
Going further, a 4th order equivariant model for example could learn highly specific chemical rules 
by itself relating to quadruples of atoms, such as ``when do these four atoms form a peptide group?''. 
Higher order relation learning would work similarly. 

Interestingly, permutation equivariant generative models are also related to the theory of finite 
exchangeable sequences, which we plan to explore in future work.  

% \clearpage 
\bibliography{SnNN}
\bibliographystyle{icml2020}

\onecolumn
\section{Appendix A:~ Supporting propositions}

\begin{prop}\label{prop: first order map}
%Let \m{S} be the quotient space \m{\Sn/\Snm}. Then a complete set of coset representatives 
%for \m{S} is provided by the set of permutations \m{\cbrN{(1,n),(2,n),\ldots,(n,n)}}.  
%\[f=\brN{}\]
%Given any first order \m{\Sn}--equivariant vector \m{f=\brN{\sseq{f}{n}}}, let the corresponding 
%element of \m{\Sn/\Snm} be  
%\[\actf{}(\mu)=f_{\mu(n)}\qqquad \mu\tin\Sn.\]
%Similarly, assume that the quotient space element corresponding to the transformed vector 
%\sm{f^\sigma=(f_{\sigma^{-1}(1)},\ldots,f_{\sigma^{-1}(n)})} ~\mbox{}~~is~\mbox{} %, the corresponding quotient space element is  
%\sm{\actf{}^\sigma(\mu)=f^\sigma_{\mu(n)}}. Then~\mbox{} \sm{\actf{}^\sigma(\mu)=\actf{}(\sigma^{-1}\mu)}. 
Let \m{f\tin\RR^n} be any vector on which \m{\Sn} acts by the first order permutation action 
\[f\overset{\sigma}{\longmapsto} f^\sigma\qquad \text{with}\qquad 
%\br{v_1,\ldots,v_n}\overset{\sigma}{\longmapsto}  
f^\sigma=\brN{f_{\sigma^{-1}(1)},\ldots,f_{\sigma^{-1}(n)}}\qqquad \sigma\tin\Sn.\] 
Let us associate to \m{f} the function \m{\mathfrak{f}\colon\Sn/\Snm\to\RR} defined 
\begin{equation}\label{eq: first order embedding}
\mathfrak{f}(\mu)=f_{\mu(n)} \qqquad \qqquad \mu\tin\Rcal, 
\end{equation}
where \m{\Rcal} is a complete set of \m{\Sn/\Snm} coset representatives. 
Then, the mapping \m{f\mapsto\mathfrak{f}} is bijective. 
%Then \m{\mathfrak{f}} 
Moreover, under permutations \sm{\mathfrak{f}\mapsto\mathfrak{f}^\sigma} with 
\sm{\mathfrak{f}^\sigma(\mu)=\mathfrak{f}(\sigma^{-1}\mu)}.  
\end{prop}
\begin{pf}
Recall that \m{\Sn/\Snm} is defined as the collection of left cosets \sm{\cbrN{\mu\,\Snm}_{\mu}} 
with \sm{\mu\ts\Snm\!:=\!\setofN{\mu\nu}{\nu\tin\Snm}\subset \Sn}. 
Any member of a given coset can be used to serve as the representative of that coset. 
Therefore, we first need to verify that \rf{eq: first order embedding} is well defined, 
i.e., that \m{\mu(n)=(\mu\nu)(n)} for any \m{\nu\tin\Snm}. 
Since \m{(\mu\nu)(n)=\mu(\nu(n))} and \m{\nu} fixes \m{n}, this is clearly true.   

%To show that the mapping \m{f\mapsto\mathfrak{f}} is bijective it is sufficient to consider that (a) 
Now, for any \m{i\tin \cbrN{\oneton{n}}} there is exactly one coset \m{\mu\ts\Snm} such that 
\m{\mu(n)\<=i} Therefore, the mapping \m{f\mapsto\mathfrak{f}} is bijective. 
Finally, to show that \m{\mathfrak{f}} transforms correctly consider that 
\m{\mathfrak{f}^\sigma(\mu):=(f^\sigma)_{\mu(n)}=f_{\sigma^{-1}(\mu(n))}=f_{(\sigma^{-1}\mu)(n)}=
\mathfrak{f}(\sigma^{-1}\mu)}. 
\end{pf}
\smallskip

The second order case is analogous with the only added complication that to ensure bijectivity we 
need to require 
%make the mapping from matrices to functions on \m{\Sn/\Snmm} bijective we need to ensure 
that the activation as a matrix be symmetric and have zero diagonal. 
For the adjacency matrices of simple graphs these conditions are satisfied automatically. 
\bigskip

\begin{prop}\label{prop: second order map}
Let \m{F\tin\RR^{n\times n}} be a matrix with zero diagonal 
on which \m{\Sn} acts by the second order permutation action 
\[F\overset{\sigma}{\longmapsto} F^\sigma\qquad \text{with}\qquad 
F^\sigma_{i,j}=F_{\sigma^{-1}(i),\ts \sigma^{-1}(j)} \qqquad \sigma\tin\Sn.\] 
Let us associate to \m{F} the function \m{\mathfrak{f}\colon\Sn/\Snmm\to\RR} defined 
\begin{equation}\label{eq: second order embedding}
\mathfrak{f}(\mu)=F_{\mu(n),\,\mu(n-1)} \qqquad \mu\tin\Rcal,
\end{equation}
where \m{\Rcal} is a complete set of \m{\Sn/\Snm} coset representatives. 
Then, the mapping \m{F\mapsto\mathfrak{f}} is bijective. 
%Then \m{\mathfrak{f}} 
Moreover, under permutations \sm{\mathfrak{f}\mapsto\mathfrak{f}^\sigma} with 
\m{\mathfrak{f}^\sigma(\mu)=\mathfrak{f}(\sigma^{-1}\mu)}.  
\end{prop}
\begin{pf}
The proof is analogous to the first order case. 
First, \m{\Sn/\Snmm} is the set of left cosets of the form 
\sm{\mu\ts\Snmm\!:=\!\setofN{\mu\nu}{\nu\tin\Snmm}}. 
Since any \m{\nu\tin\Snmm} fixes \m{\cbrN{n,n-1}}, for any \m{\mu\nu\tin\mu\ts\Snmm},  
\m{\mu\nu(n)=\mu(n)} and \m{\mu\nu(n\<-1)=\mu(n\<-1)}. Therefore \m{\mathfrak{f}} is well defined.  

It is also easy to see that for any \m{(i,j)} pair with \m{i,j\tin\cbrN{\oneton{n}}} and \m{i\<\neq j}, then  
there is exactly one \m{\mu\,\Snmm} coset satisfying \m{\mu(n)\<=i} and \m{\mu(n\<-1)\<=j}. 
Finally, to show that \m{\mathfrak{f}} transforms correctly, 
\sm{\mathfrak{f}^\sigma(\mu)=(F^\sigma)_{\mu(n),\ts\mu(n-1)}=
F_{\sigma^{-1}(\mu(n)),\ts\sigma^{-1}(\mu(n-1))}=F_{(\sigma^{-1}\mu)(n),\,(\sigma^{-1}\mu)(n-1)}=
\mathfrak{f}(\sigma^{-1}\mu)}. 
\end{pf}

%The \m{k}'th order case is a direct generalization of the first and second order ones. Its 
%proof is omitted because it is analogous.  
%\bigskip

\begin{prop}\label{prop: kth order map}
Let \m{F\tin\RR^{n\times \ldots\times n}} be a \m{k}'th order tensor 
%that is symmetric in all its indices and 
satisfying \m{F_{\sseq{i}{k}}=0} unless \m{\sseq{i}{k}} are all distinct.  
Assume that \m{\Sn} acts on \m{F} by the \m{k}'th order permutation action 
\[F\overset{\sigma}{\longmapsto} F^\sigma\qquad \text{with}\qquad 
F^\sigma_{\sseq{i}{k}}=F_{\sigma^{-1}(i_1),\ldots,\sigma^{-1}(i_k)} \qqquad \sigma\tin\Sn.\] 
Let us associate to \m{F} the function \m{\mathfrak{f}\colon\Sn/\Sbb_{n-k}\to\RR} defined 
\begin{equation}\label{eq: kth order embedding}
\mathfrak{f}(\mu)=F_{\mu(n),\ldots,\mu(n-k+1)} \qqquad \mu\tin\Rcal,
\end{equation}
where \m{\Rcal} is a complete set of \m{\Sn/\Sbb_{n-k}} coset representatives. 
Then, the mapping \m{F\mapsto\mathfrak{f}} is bijective. 
%Then \m{\mathfrak{f}} 
Moreover, under permutations \sm{\mathfrak{f}\mapsto\mathfrak{f}^\sigma} with 
\m{\mathfrak{f}^\sigma(\mu)=\mathfrak{f}(\sigma^{-1}\mu)}.  
\end{prop}

\textbf{Proof.}~ 
%\begin{pf}
Analogous to the first and and second order cases. 
%\end{pf}
%The significance of Propositions \ref{prop: first order map}--\ref{prop: kth order map} are that 
%they allow us to use the results of %\citep{}.

\ignore{
be of the form  
\[\phi_\ell(\actn{\ell-1})=\actn{\ell-1}\ast \chi_\ell,\]
where \m{\ast} denotes a generalized notion of convolution of functions on \m{G}. In the 
context of \m{\Sn}--equivariance we have the following result. 

\begin{prop}
Let \m{\Ncal} be a permutation equivariant convolutional neural network where 
\end{prop}
}

\section{Appendix B:~ Proofs}

\begin{pfof}{Theorem 1} %\ref{thm: convo1}}
Let \m{\ffin} and \m{\ffout} be the quotient space functions corresponding to \m{\fin} and \m{\fout}. 
Then, by Theorem 1 of (Kondor \& Trivedi, 2018), 
%\citep{KondorTrivedi2018}, %\m{\ffout=\xi(\ffin\ast \chi)} 
\[\ffout(\mu)=\xi\brbigg{\:\sum_{\nu\in\Sn} \ffin(\mu\nu^{-1})\,\chi(\nu)}\]
for some appropriate 
pointwise nonlinearity \m{\xi} and some (learnable) filter \m{\chi\colon \Snm\backslash\Sn/\Snm\to\RR}. 
Alternatively, mapping \m{\ffout} back to vector form, we can write 
\[\fout_i=\xi(\fpre_i)\qqquad \qqquad 
\fpre_i=\sum_{\nu\in\Sn} \ffin(\mu_i\nu^{-1})\,\chi(\nu),
\]
where \m{\mu_i} denote the representative of the coset that maps \m{n\mapsto i}, 

%Following standard conventions, 
Let \m{e} denote the identity element of \m{\Sn} and 
\m{\tau_{n,n\<-1}\tin\Sn} denote the element that swaps \m{n} and \m{n\<-1}. 
There are only two \m{\Snm\backslash \Sn/\Snm} cosets:
\begin{eqnarray*}
&S_0=\Snm e\,\Snm=\setof{\mu\tin\Sn}{\mu(n)=n}\\
&S_1=\Snm \tau_{n,n-1}\,\Snm=\setof{\mu\tin\Sn}{\mu(n)\neq n}.
\end{eqnarray*}
Assume that \m{\chi} takes on the value \m{\chi_0} on \m{S_0} and the 
value \m{\chi_1} on \m{S_1}. Then 
\[%\ffout(\mu)=\xi\brbigg{\:\sum_{\nu\in\Sn} \ffin(\mu\nu^{-1})\,\chi(\nu)}=
\fpre_i=
%\xi\brbigg{\:
\sum_{\nu\in S_0} \ffin(\mu_i\nu^{-1})\,\chi(\nu)+
\sum_{\nu\in S_1} \ffin(\mu_i\nu^{-1})\,\chi(\nu).
\]
For the \m{\nu\tin S_0} case note that \m{\abs{S_0}=(n-1)!} and that 
\m{\mu} and \m{\mu\nu^{-1}} will always fall in the same left \m{\Snm}-coset. Therefore,  
\[\sum_{\nu\in S_0} \ffin(\mu_i\nu^{-1})\,\chi(\nu)=
\chi_0 \sum_{\nu\in S_0} \ffin(\mu_i\nu^{-1})=
(n-1)!\: \chi_0\: \ffin(\mu_i)=(n-1)!\: \chi_0\: \fin_i.\]
For the \m{\nu\tin S_1} case note that \m{\nu} as traverses \m{S_1}, \m{\mu\nu^{-1}} will \emph{never} 
fall in the same coset as \m{\mu}, but will fall in each of the other cosets exactly 
\m{(n-1)!} times. Therefore, 
\[\sum_{\nu\in S_1} \ffin(\mu_i\nu^{-1})\,\chi(\nu)=
\chi_1 \sum_{\nu\in S_1} \ffin(\mu_i\nu^{-1})=
(n\<-1)!\: \chi_0\!\!\!\!\! \sum_{\substack{\pi\in\Sn/\Snm\\ \pi\not\in\mu_i\Snm}}\!\!\!\! \ffin(\pi)=
(n\<-1)!\: \chi_0 \sum_{j\neq i} \fin_j.
\]
%Letting \m{\mu_i} denote the representative of the coset that maps \m{n\mapsto i}, 
%\m{}
Combining the above, 
\[\fout_i=
\xi\sqbBig{(n\<-1)!\:\brBig{\chi_0\, \fin_i+\chi_1\sum_{j\neq i}\fin_j}}=
\xi\sqbBig{(n\<-1)!\:\brBig{\chi_0\, \fin_i+\chi_1\sum_{j}\fin_j-\chi_1 \fin_i}}.
\]
Setting \m{w_0=(n-1)!(\chi_0-\chi_1)} and \m{w_1=(n-1)!\:\chi_1} proves the result. 
\end{pfof}
%\bigskip

%\hline 
%\bigskip

\begin{pfof}{Theorem 2} %\ref{thm: convo2}}
Let \m{\ffin}, \m{\ffout}, \m{\tau_{i,j}} and \m{\chi} be as in the proof of Theorem 1. %\ref{thm: convo1}. 
Now, however, \m{f} is indexed by two indices, \m{i,j\tin\cbrN{\oneton{n}}} and \m{i\<\neq j}. 
Correspondingly, we let \m{\mu_i} denote the representative of the \m{\Snmm}--coset consisting 
of permutations that take \m{n\<\mapsto i} and \m{n\<-1\<\mapsto j}. 
Similarly to the proof of Theorem 1, we set %\ref{thm: convo1}, we set 
\[\fout_{i,j}=\xi(\fpre_{i,j})\qqquad \qqquad 
\fpre_{i,j}=\sum_{\nu\in\Sn} \ffin(\mu_{i,j}\nu^{-1})\,\chi(\nu),
\vspace{-6pt}
\]
where now  \m{\chi\colon \Snmm\backslash\Sn/\Snmm\to\RR}.
 
There are a rotal of seven  \m{\Snmm\backslash\Sn/\Snmm\to\RR} cosets:
\begin{eqnarray*}
&&S_0=\Snmm\, e\,\Snmm=\setof{\mu\tin\Sn}{\mu(n)\<=n,~ \mu(n-1)\<=n\<-1}\\
&&S_1=\Snmm\, \tau_{n,n-1}\,\Snmm=\setof{\mu\tin\Sn}{\mu(n)\<=n\<-1,~ \mu(n-1)\<=n}\\
&&S_2=\Snm\,\tau_{n-1,n-2}\,\Snm=\setof{\mu\tin\Sn}{\mu(n)\<=n,~ \mu(n-1)\tin\cbrN{1,\ldots,n-2}},\\
&&S_3=\Snm\,\tau_{n,n-1}\,\tau_{n-1,n-2}\,\Snm=\setof{\mu\tin\Sn}{\mu(n)\<=n\<-1,~ \mu(n-1)\tin\cbrN{1,\ldots,n-2}},\\
&&S_4=\Snm\,\tau_{n,n-2}\,\Snm=\setof{\mu\tin\Sn}{\mu(n)\tin\cbrN{1,\ldots,n-2},~ \mu(n-1)\<=n\<-1},\\
&&S_5=\Snm\,\tau_{n,n-1}\,\tau_{n,n-2}\,\Snm=\setof{\mu\tin\Sn}{\mu(n)\tin\cbrN{1,\ldots,n-2},~ \mu(n-1)\<=n},\\
&&S_6=\Snm\,\tau_{n,n-2}\,\tau_{n-1,n-3}\,\Snm=
\setof{\mu\tin\Sn}{\mu(n)\tin\cbrN{1,\ldots,n-2},~ \mu(n\<-1)\tin\cbrN{1,\ldots,n-2}}.
\end{eqnarray*}
Assuming that \m{\chi} takes on the values \m{\chi_0,\ldots,\chi_6} on these seven cosets, 
\[
\fpre_{i,j}=%\sum_{\nu\in\} \ffin(\mu\nu^{-1})\,\chi(\nu)}=
\sum_{p=0}^{6} \:\chi_p \! \underbrace{\sum_{\nu\in S_p} \ffin(\mu_{i,j} \nu^{-1})}_{h_{i,j}^p}. %=:
%\sum_{p=0}^{6} \:\chi_p\, h^p_{i,j}. % \! \sum_{\nu\in S_p} \ffin(\mu_{i,j} \nu^{-1}).
\vspace{-6pt}
\]
Now we analyze the \m{p=0,1,\ldots,6} cases separately. 

\begin{enumerate}[itemsep=0pt, topsep=0pt, leftmargin=18pt, labelsep=6pt, label={\m{\circ}}]
\item 
In the \m{p\<=0} case \m{\abs{S_0}\<=(n\<-2)!}~\mbox{} and~\mbox{} \m{\mu_{i,j}\nu^{-1}\!\tin\mu_{i,j}\ts\Snm}, 
therefore 
\[h_{i,j}^0=\sum_{\nu\in S_0} \ffin(\mu_{i,j}\nu^{-1})=(n-2)!\: \fin_{i,j}.\]
\item 
In the \m{p\<=1} case \m{\abs{S_1}\<=(n\<-2)!}~\mbox{} and~\mbox{} \m{\mu_{i,j}\nu^{-1}\!\tin\mu_{j,i}\ts\Snm}, 
therefore 
\[h_{i,j}^1=\sum_{\nu\in S_1} \ffin(\mu_{i,j}\nu^{-1})=(n-2)!\: \fin_{j,i}.\]
\item 
In the \m{p\<=2} case, as \m{\nu} traverses \m{S_2},~\mbox{} \m{\mu_{i,j}\nu^{-1}} will hit each 
\m{\mu_{i,k}\ts\Snmm} coset with \m{k\!\not\in\!\cbrN{i,j}} exactly \m{(n\<-2)!} times, therefore 
\[h_{i,j}^2=\sum_{\nu\in S_2} \ffin(\mu_{i,j}\nu^{-1})=(n-2)!\: \sum_{k\not\in\cbr{i,j}} \fin_{i,k}.\]
\item 
In the \m{p\<=3} case, as \m{\nu} traverses \m{S_3},~\mbox{} \m{\mu_{i,j}\nu^{-1}} will hit each 
\m{\mu_{k,i}\ts\Snmm} coset with \m{k\!\not\in\!\cbrN{i,j}} exactly \m{(n\<-2)!} times, therefore 
\[h_{i,j}^3=\sum_{\nu\in S_3} \ffin(\mu_{i,j}\nu^{-1})=(n-2)!\: \sum_{k\not\in\cbr{i,j}} \fin_{k,i}.\]
\item 
In the \m{p\<=4} case, as \m{\nu} traverses \m{S_4},~\mbox{} \m{\mu_{i,j}\nu^{-1}} will hit each 
\m{\mu_{k,j}\ts\Snmm} coset with \m{k\!\not\in\!\cbrN{i,j}} exactly \m{(n\<-2)!} times, therefore 
\[h_{i,j}^4=\sum_{\nu\in S_4} \ffin(\mu_{i,j}\nu^{-1})=(n-2)!\: \sum_{k\not\in\cbr{i,j}} \fin_{k,j}.\]
\item 
In the \m{p\<=5} case, as \m{\nu} traverses \m{S_5},~\mbox{} \m{\mu_{i,j}\nu^{-1}} will hit each 
\m{\mu_{j,k}\ts\Snmm} coset with \m{k\!\not\in\!\cbrN{i,j}} exactly \m{(n\<-2)!} times, therefore 
\[h_{i,j}^5=\sum_{\nu\in S_5} \ffin(\mu_{i,j}\nu^{-1})=(n-2)!\: \sum_{k\not\in\cbr{i,j}} \fin_{j,k}.\]
\item 
In the \m{p\<=6} case, as \m{\nu} traverses \m{S_6},~\mbox{} \m{\mu_{i,j}\nu^{-1}} will hit each 
\m{\mu_{k,l}\ts\Snmm} coset with \m{k\!\not\in\!\cbrN{i,j}} and \m{l\!\not\in\!\cbrN{i,j}} exactly \m{(n\<-2)!} times, therefore 
\[h_{i,j}^6=\sum_{\nu\in S_6} \ffin(\mu_{i,j}\nu^{-1})=(n-2)!\: \sum_{k,l\not\in\cbr{i,j}} \fin_{k,l}.\]
\end{enumerate}
 
Summing each of the above terms gives 
\begin{eqnarray*}
\fout_{i,j}=
\xi\brBig{(n\<-2)!\:\brbig{\chi_0\ts\fin_{i,j}+\chi_1\ts\fin_{j,i}+\chi_2(\fin_{i,\ast}-\fin_{i,j})
+\chi_3(\fin_{\ast,i}-\fin_{j,i})+\chi_4(\fin_{\ast,j}-\fin_{i,j})+\\
\chi_5(\fin_{j,\ast}-\fin_{j,i})+
\chi_6(\fin_{\ast,\ast}-\fin_{i,*}-\fin_{*,j}+\fin_{i,j}-\fin_{j,*}-\fin_{*,i}+\fin_{j,i})
}}.
\end{eqnarray*}
The results follows by setting \vspace{-10pt}
\begin{eqnarray*}
&&w_0=\chi_0-\chi_2-\chi_4+\chi_6\\
&&w_1=\chi_1-\chi_3-\chi_5+\chi_6\\
&&w_2=\chi_2-\chi_6\\
&&w_3=\chi_3-\chi_6\\
&&w_4=\chi_4-\chi_6\\
&&w_5=\chi_5-\chi_6\\
&&w_6=\chi_6.
\end{eqnarray*}
%where \m{\fin_{i,\ast}=\sum_k \fin_{i,k}},~\mbox{} \m{\fin_{\ast,j}=\sum_k \fin_{k,j}},~\mbox{} 
%and \m{\fin_{\ast,\ast}=\sum_{k,l} \fin_{k,l}}.
%\[
%\fout_{i,j}=\xi\brBig{w_0 \fin_{i,j}+w_1 \fin_{j,i}+w_2 \fin_{i,\ast} +w_3 \fin_{\ast,i}+
%w_4 \fin_{\ast,j}+w_5 \fin_{j,\ast}+w_6\fin_{\ast,\ast}}. 
%\]
\vspace{-12pt}\mbox{}
\end{pfof}

\begin{pfof}{Corollary 4.} 
The general theory of harmonic analysis on \m{\Sn} tells us that the off-diagonal part of \m{\fin} 
transforms as a representation of type \m{(1,2,1,1)} whereas the diagonal part transforms as a representation 
of type \m{(1,1)}. Putting these together, the overall type of \m{\fin} is \m{(2,3,1,1)}. 
Therefore, the Fourier transform of \m{\fin} consists of four matrices \m{F_1,F_2,F_3,F_4}, and the corresponding \m{W_i} 
weight matrices are of size \m{2\<\times 2}, \m{3\<\times 3}, \m{1\times 1} and \m{1\times 1}, respectively. 

By Theorem 3, \m{\fout_{i,j}=\xi(\tilde f_{i,j})}, where the Fourier transform of 
\m{\tilde f_{i,j}} is \m{(F_1 W_1, F_2 W_2, F_3 W_3, F_4 W_4)}. 
Therefore, for a given \m{\fin} any given component \m{\tilde f_{i,j}} of \m{\tilde f} lives in a 15 
dimensional space parametrized by the 15 entries of the \m{W_i} weight matrices. 
It is easy to see that the various polynomials \m{\fin_{i,j}, \fin_{j,i}, \fin_{i,\ast}}, etc.~appearing in 
Corollary 4 are all equivariant and they are linearly independent of each other. 
Since there are exactly 15 of them, so the expression for \m{\fout_{i,j}} appearing in the 
Corollary is indeed the most general possible.
\end{pfof}
\section{Appendix C:~ Third order Layer}

Here, we describe the form of the third-order layer.
The functional form for a layer that maps a third order \m{\Sn}--equivariant activation \m{\fin\tin\RR^{n\times n \times n}} 
to another third-order \m{\Sn}--equivariant activation \m{\fout\in\RR^{n\times n \times n}}
can be derived in a manner similar to Corollary~4.
\subsection{Counting the number of parameters}\label{ssec:third_order_num_params}
Just as the second order activation could be decomposed into a second order off-diagonal component and a first order diagonal,
here we decompose a third-order tensor $f_{ijk}$ into five types of elements:
\begin{enumerate}
    \item Elements where $i\neq j \neq k$
    \item Elements where $i=j \neq k$,
    \item Elements where $i\neq j =k$
    \item Elements where $i=k \neq j$
    \item Elements where $i=j = k$
\end{enumerate}
Each type of item is band-limited in Fourier space, consisting only of Young tableaus $\left[ n-3, 1, 1, 1 \right]$ or greater.
Specifically, elements of type 1) are representations of type $\left(1, 3, 3, 3, 1, 2, 1\right)$,
elements of type 2-4) are representations of type $\left(1, 2, 1, 1, 0, 0, 0\right)$,
and elements of type 5) are representations of type $\left(1, 1, 0, 0, 0, 0, 0\right)$.
By the same arguments as before, this means the number of free parameters in the convolutional layer is given by
\begin{equation*}
    \left( 1 + 3 \times 1 + 1 \right)^2 + \left( 3 + 3 \times 2 + 1 \right)^2 + \left( 3 + 3 \times 1  \right)^2 + \left( 3 + 3 \times 1\right)^2
     + 1^2  + 2^2 + 1^2= 203
\end{equation*}
\subsection{Constructing the layer}
Due to the large number of parameters, we will not explicitly explicitly give the form of the layer.
However, we will describe the construction of the layer algorithmically.
We begin by writing the possible contractions that can be the added to $f_{ijk}$.
\begin{enumerate}
    \item Contractions giving third order tensors:
        \begin{enumerate}
            \item Keep the original value $f_{ijk}$.  This can be done only one way.
        \end{enumerate}
    \item Contractions giving second order tensors:
        \begin{enumerate}
            \item Sum over a single index, e.g. $\sum_i f_{ijk}$ or $\sum_j f_{ijk}$
                This gives three tensors: one for each index. 
            \item Set one index equal to a second one, e.g. $f_{iik}$, $f_{ijj}$.
                This gives three tensors: one for each unchanged index.
        \end{enumerate}
    \item Contractions giving first order tensors:
        \begin{enumerate}
            \item Sum over two indices, e.g. $\sum_{ij} f_{ijk}$ or $\sum_j f_{ijk}$
                This gives three tensors: one for each index that isn't summed over. 
            \item Set one index equal to a second one and then sum over the third, e.g. $\sum_k f_{iik}$.
                This gives three tensors, each corresponding to the summed index.
            \item Set one index equal to a second one and then sum over it, e.g. $\sum_i f_{iik}$, $f_{ijj}$.
                This gives three tensors, each corresponding to the untouched index.
            \item Set all indexes equal to each other, e.g. $f_{iii}$, $f_{jjj}$, $f_{kkk}$.
                This gives one tensor.
        \end{enumerate}
    \item Contractions giving zeroth order tensors:
        \begin{enumerate}
            \item Sum over all indices, $\sum_{ijk} f_{ijk}$
                This gives one tensor.
            \item Set one index equal to a second one, sum over it, and then sum over the third, e.g. $\sum_{i, k} f_{iik}$.
                This gives three tensors, each corresponding to the index that is not set equal before summation.
            \item Set all indexes equal to each other and then sum over them e.g. $\sum_i f_{iii}$
                This gives three tensors.
        \end{enumerate}
\end{enumerate}
In summary, the contractions yield 1 third-order tensor, 6 second order tensors, 10 first order tensors, and 5 zeroth order tensors.
We now add the elements of the contracted tensors together to form $\fout$.  As we can permute the indices of the contracted tensors without breaking equivariance, we consider all possible permutations as well.
For instance,  when $i \neq j \neq k$, we have  
\begin{equation*}
    \fout_{ijk} = w_0 \fin_{ijk} + w_1 \fin_{jik}  + w_2 \fin_{ikj} \ldots  + w_6 A_{ij}  + w_7 A_{ji} + w_8 A_{ik} + \ldots + w_{42} B_i + w_{43} B_j \ldots
\end{equation*}
where $A$ is a generic second-order contraction and $B$ is a generic first-order contraction.
This means that, for a generic element in $\fout_{ijk}$, we have 6 terms from a single third order contraction, 6 terms from a single second order contraction, 3 terms from a single first order contraction, and 1 term from a zeroth contraction.

However, as in Corollary 4, elements where two or more indices match have additional flexibility, as they can be treated as linear combinations with separate, lower order tensors.
Consequently, we can add additional contractions to those terms of equal or lower order.
The total number of terms that contribute to each part of $\fout$ is summarized in Table~\ref{tbl:components}.
\begin{table}[t]
    \caption{\label{tbl:components} How many terms in a given component of $\fout$ are from a given contraction.}
\begin{center}
\begin{tabular}{||l | c | c | c | c | r ||}
	\hline
	$\fout$ Component & 3rd order Contr. & 2nd order Contr. & 1st order Contr.  & 0th order Contr.  &  Total \\ 
	\hline\hline
    $> 0$ indices shared &  6 $\times$ 1 & 6 $\times$ 6 & 3 $\times$ 10 & 1 $\times$ 5 & 77 \\
	\hline
    $> 1$ index shared &  0 $\times$ 1 & 2 $\times$ 6 & 2 $\times$ 10 & 1 $\times$ 5 & 37 \\
	\hline
    $> 2$ indices shared &  0 $\times$ 1 & 0 $\times$ 6 & 1 $\times$ 10 & 1 $\times$ 5 & 15\\
	\hline
\end{tabular}
\end{center}
\end{table}
This gives a total of $77 + 3 \times 37  + 15 = 203$ linear weights, which matches the results of Subsection~\ref{ssec:third_order_num_params}.

\section{Effect of Loss Function on Molecular Generation}

% To further explore the effect of the latent distribution on the network, 
% we trained two additional variational autoencoders on the ZINC dataset.
% However, in one case the Kullbeck-Leibler divergence term in the VAE loss function was disabled ($\beta=0$).
When training the $Sn$-Conv VAE, we disabled the VAE loss function ($\beta=0$).
This removes the penalty constraining the amortization distribution from 
being close to the prior.
Initially, we might expect that turning off the KL divergence would increase reconstruction accuracy as the model has less constraints in the latent layer.
At the same time, it would decrease molecular validity, as the amortized distribution would drift further from the $N(0, 1)$ prior.
% In both cases a latent layer of 20-dimensional space per node was used.
% A larger latent space was used in an attempt to ensure that no artificial constraints on the latent distribution were placed by its limited dimensionality.

\begin{table}[b]
\caption{\label{tbl:zinc_2} ZINC results with latent layer of dimension 20.}
\begin{center}
\small{
\begin{tabular}{||l | c | c | c||}
	\hline
	Method & Accuracy & Validity & Unique\\ 
	\hline
    2nd order $\Sn$-Conv VAE w. KL& {87.1\%} & {93.\%} &  7.3\%\\
	\hline
    2nd order $\Sn$-Conv VAE w.o. KL& {98.4\%} & {33.4\%} & 82.7\%\\
	\hline
\end{tabular}
}
\end{center}
\end{table}
At first glancer, our results shown in Table~\ref{tbl:zinc_2}, would support this hypothesis.  
Indeed, we see higher reconstruction accuracy, but lower validity, when we disable the KL divergence term. 
However, closer inspection shows that the high validity of the 2nd-order VAE is an artifact: the model is merely choosing to predict multiple small molecules with five atoms or less.  
This is reflected by the considerably lower uniqueness of the valid molecules produced when the KL divergence is used.
% This is a considerably lower fraction than the 58\% of unique molecules produced by the architecture in the main paper.
Indeed, if the KL divergence included, the architecture only produces 8 valid molecules with more than 10 toms in 5000 attempts.
Omitting the KL divergence, in contrast, produces 355.

One explanation for these results might be mode-collapse.  
However, the fact that the architecture with KL divergence can still successfully reconstruct encoded molecules, suggests this is not the case.
Rather, these results may point towards the limits of simple, IID latent distributions, as forcing the distribution seems to worsen the generative results of the network. 
This suggests that the construction of networks that leverage more complex, exchangeable distributions may lead to improved results.

\end{document}